\documentclass[10pt,twocolumn,letterpaper]{article}

\usepackage[pagenumbers]{cvpr} %

\usepackage[dvipsnames]{xcolor}
\usepackage{amsfonts}
\usepackage{bm}
\usepackage{amsmath}
\usepackage{mathtools}
\usepackage{multirow}
\usepackage{booktabs}
\usepackage{caption}
\usepackage{enumitem}

\newcommand\blfootnote[1]{%
  \begingroup
  \renewcommand\thefootnote{}\footnote{#1}%
  \addtocounter{footnote}{-1}%
  \endgroup
}

\renewcommand{\paragraph}[1]{{\vspace{1mm}\noindent \bf #1}.}

\newcommand{\btau}{\boldsymbol{\tau}}
\newcommand{\bmu}{\boldsymbol{\mu}}
\newcommand{\beps}{\boldsymbol{\epsilon}}
\newcommand{\bpsi}{\boldsymbol{\Psi}}
\newcommand{\bes}{\mathbf{s}}
\newcommand{\ba}{\mathbf{a}}
\newcommand{\bx}{\mathbf{x}}
\newcommand{\bc}{\mathbf{c}}
\newcommand{\bp}{\mathbf{p}}
\newcommand{\bz}{\mathbf{z}}

\newcommand{\map}{\boldsymbol{\mathcal{M}}}
\newcommand{\reals}{\mathbb{R}}
\newcommand{\normal}{\mathcal{N}}
\newcommand{\guide}{\mathcal{J}}

\newcommand{\name}{{{TRACE}}\xspace}
\newcommand{\animname}{{{PACER}}\xspace}

\usepackage{graphicx}
\usepackage{amsmath}
\usepackage{amssymb}
\usepackage{booktabs}

\usepackage[pagebackref,breaklinks,colorlinks]{hyperref}

\usepackage[capitalize]{cleveref}
\crefname{section}{Sec.}{Secs.}
\Crefname{section}{Section}{Sections}
\Crefname{table}{Table}{Tables}
\crefname{table}{Tab.}{Tabs.}

\def\cvprPaperID{4784} %
\def\confName{CVPR}
\def\confYear{2023}

\begin{document}

\title{Trace and Pace:  Controllable Pedestrian Animation \\ via Guided Trajectory Diffusion}

\author{
\vspace{1mm}
Davis Rempe$^{*,1,2}$\qquad Zhengyi Luo$^{*,1,3}$\qquad Xue Bin Peng$^{1,4}$ \qquad Ye Yuan$^{1}$ \qquad Kris Kitani$^{3}$ \\ \vspace{1mm}
Karsten Kreis$^{1}$ \qquad Sanja Fidler$^{1,5,6}$ \qquad Or Litany$^{1}$ \\ %
\small $^1$NVIDIA\qquad $^2$Stanford University \qquad $^3$Carnegie Mellon University  \qquad $^4$Simon Fraser University \\ \small $^5$University of Toronto \qquad $^6$Vector Institute
}

\twocolumn[{
\renewcommand\twocolumn[1][]{#1}%
\maketitle
\vspace{-0.4in}
\begin{center}
    \centering
    \includegraphics[width=\textwidth]{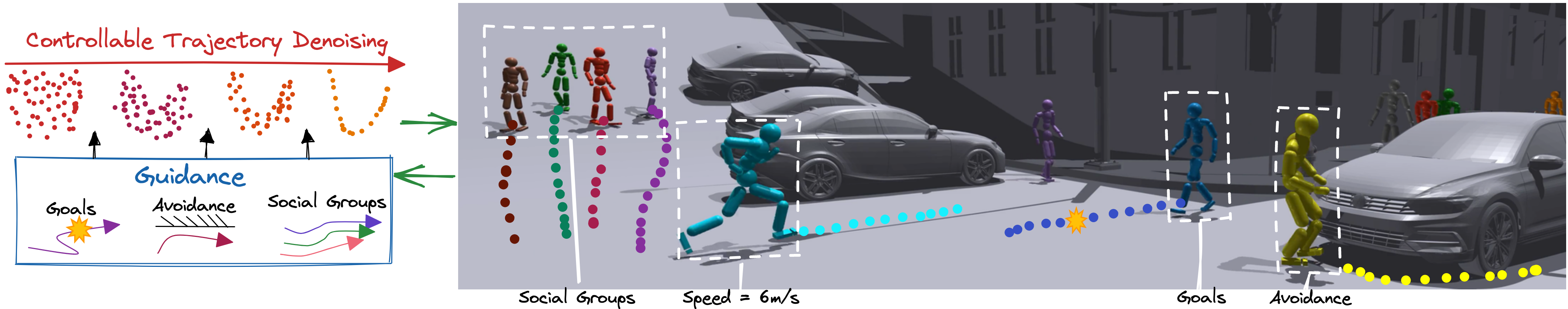}
    \vspace{-7mm}
    \captionof{figure}{(Left) We propose \name, a trajectory diffusion model that enables user control through test-time guidance. (Right) Generated trajectories are passed to a novel physics-based humanoid controller (\animname), forming a closed-loop pedestrian animation system.}
    \label{fig:teaser}
\end{center}%
}]
\maketitle

\begin{abstract}

We introduce a method for generating realistic pedestrian trajectories and full-body animations that can be controlled to meet user-defined goals. We draw on recent advances in guided diffusion modeling to achieve test-time controllability of trajectories, which is normally only associated with rule-based systems. Our guided diffusion model allows users to constrain trajectories through target waypoints, speed, and specified social groups while accounting for the surrounding environment context. This trajectory diffusion model is integrated with a novel physics-based humanoid controller to form a closed-loop, full-body pedestrian animation system capable of placing large crowds in a simulated environment with varying terrains. We further propose utilizing the value function learned during RL training of the animation controller to guide diffusion to produce trajectories better suited for particular scenarios such as collision avoidance and traversing uneven terrain.
Video results are available on the \href{https://nv-tlabs.github.io/trace-pace}{\texttt{project page}}.

\end{abstract}
\vspace{-3mm}
\section{Introduction}
\label{sec:intro}
\blfootnote{$^*$Equal contribution}
Synthesizing high-level human behavior, in the form of 2D positional trajectories, is at the core of modeling pedestrians for applications like autonomous vehicles, urban planning, and architectural design.
An important feature of such synthesis is \emph{controllability} -- generating trajectories that meet user-defined objectives, edits, or constraints.
For example, a user may place specific waypoints for characters to follow, specify social groups for pedestrians to travel in, or define a social distance to maintain.

Attaining controllability is straightforward for algorithmic or rule-based models of human behavior, since they have built-in objectives. 
In the simplest case, human trajectories can be determined by the shortest paths between control points~\cite{hassan2021stochastic}, but more sophisticated heuristics have also been developed for pedestrians~\cite{helbing1995social,berg2011reciprocal}, crowds~\cite{ren2017group,kim2014interactive}, and traffic~\cite{treiber2000congested,lopez2018microscopic}. 
Unfortunately, algorithmically generated trajectories often appear unnatural.
Learning-based approaches, on the other hand, can improve naturalness by mimicking real-world data.
These methods often focus on short-term trajectory prediction using a single forward pass of a neural network~\cite{gupta2018social,yuan2021agentformer,salzmann2020trajectron,alahi2016social}.
However, the ability to \emph{control} these models is limited to sampling from an output trajectory distribution~\cite{mangalam2021goals,xu2022bits} or using an expensive latent space traversal~\cite{rempe2022strive}.
As a result, learning-based methods can predict implausible motions such as collisions with obstacles or between pedestrians.
This motivates another notion of \emph{controllability} -- maintaining realistic trajectories during agent-agent and agent-environment interactions.

In this work, we are particularly interested in using controllable pedestrian trajectory models for character animation. We envision a simple interface where a user provides high-level objectives, such as waypoints and social groups, and a system converts them to \textit{physics-based} full-body human motion. Compared to existing kinematic motion models~\cite{priisalu2020semantic, holden2017phase, ling2020character}, physics-based methods have the potential to produce high-quality motion with realistic subtle behaviors during transitions, obstacle avoidance, traversing uneven terrains, \etc. Although there exist physics-based animation models~\cite{peng2017deeploco, Haworth2020DeepIO, peng2021amp,peng2022ase,ling2020character,won2022physics}, controlling their behavior requires using task-specific planners that need to be re-trained for new tasks, terrains, and character body shapes.

 We develop a generative model of trajectories that is data driven, controllable, and tightly integrated with a physics-based animation system for full-body pedestrian simulation (\cref{fig:teaser}). 
Our method enables generating pedestrian trajectories that are realistic and amenable to user-defined objectives at test time. %
We use this trajectory generator as a planner for a physics-based pedestrian controller, resulting in a closed-loop controllable pedestrian animation system.

For trajectory generation, we introduce a \textbf{TRA}jectory Diffusion Model for \textbf{C}ontrollable P\textbf{E}destrians (\name). Inspired by recent successes in generating trajectories through denoising~\cite{gu2022stochastic,janner2022diffuser,zhong2022ctg}, 
\name generates the future trajectory for each pedestrian in a scene and accounts for the surrounding context through a spatial grid of learned map features that is queried locally during denoising.
We leverage classifier-free sampling~\cite{ho2022classifier} to allow training on mixed annotations (\eg, with and without a semantic map), which improves controllability at test time by trading off sample diversity with compliance to conditioning.
User-controlled sampling from \name is achieved through test-time \emph{guidance}~\cite{dhariwal2021diffusion,ho2022classifier,ho2022video}, which perturbs the output at each step of denoising towards the desired objective.
We extend prior work~\cite{janner2022diffuser} by introducing several analytical loss functions for pedestrians and re-formulating trajectory guidance to operate on clean trajectory outputs from the model~\cite{ho2022video}, improving sample quality and adherence to user objectives.

For character animation, we develop a general-purpose \textbf{P}edestrian \textbf{A}nimation \textbf{C}ontroll\textbf{ER} (\animname) capable of driving physics-simulated humanoids with diverse body types to follow trajectories from a high-level planner. We focus on (1) motion quality: \animname learns from a small motion database to create natural and realistic locomotion through adversarial motion learning \cite{peng2021amp, peng2022ase}; (2) terrain and social awareness: trained in diverse terrains with other humanoids, \animname learns to move through stairs, slopes, uneven surfaces, and to avoid obstacles and other pedestrians; (3) diverse body shapes: by training on different body types, \animname draws on years of simulation experience to control a wide range of characters; (4) compatibility with high-level planners: \animname accepts 2D waypoints and can be a plug-in model for any 2D trajectory planner. 

We demonstrate a controllable pedestrian animation system using \name as a high-level planner for \animname, the low-level animator. The planner and controller operate in a closed loop through frequent re-planning according to simulation results. We deepen their connection by guiding \name with the value function learned during RL training of \animname to improve animation quality in varying tasks.
We evaluate \name on synthetic~\cite{berg2011reciprocal} and real-world pedestrian data~\cite{pellegrini2009you,lerner2007crowds,caesar2020nuscenes}, demonstrating its flexibility to user-specified and plausibility objectives while synthesizing realistic motion. 
Furthermore, we show that our animation system is capable and robust with a variety of tasks, terrains, and characters.
In summary, we contribute 
(1) a diffusion model for pedestrian trajectories that is readily controlled at test time through guidance, 
(2) a general-purpose pedestrian animation controller for diverse body types and terrains, and
(3) a pedestrian animation system that integrates the two to drive simulated characters in a controllable way.

\vspace{-1mm}
\section{Related Work}
\label{sec:relwork}
\vspace{-1mm}
\paragraph{Pedestrian Trajectory Prediction}
Modeling high-level pedestrian behavior has been extensively studied in the context of motion prediction (forecasting).
Approaches range from physics and planning-based~\cite{helbing1995social,van2008reciprocal,helbing2000simulating} to recent learned methods~\cite{yuan2021agentformer,salzmann2020trajectron,alahi2016social,chai2019multipath,lee2017desire}.
We refer the reader to the thorough survey by Rudenko \etal ~\cite{rudenko2020human} for an overview, and focus this discussion on controllability.
Most forecasting work is motivated by planning for autonomous vehicles (AVs) or social robots~\cite{gupta2018social} rather than \textit{controllability} or longer-term synthesis.
Rule-based models for pedestrians~\cite{berg2011reciprocal,ren2017group,kim2014interactive} and vehicle traffic~\cite{treiber2000congested,lopez2018microscopic} can easily incorporate user constraints~\cite{lemonari2022authoring} making them amenable to control.
However, the trajectories of these approaches are not always human-like; methods have even been developed to choose the best simulation method and tune parameters to make crowd scenarios more realistic~\cite{karamouzas2018crowd}.

Data-driven methods produce human-like motions, but neural network-based approaches are difficult to explicitly \emph{control}.
Some works decompose forecasting into goal prediction followed by trajectory prediction based on goals~\cite{mangalam2021goals,dendorfer2020goal}. These models offer limited control by selecting goal locations near a target or that minimizes an objective (\eg collisions)~\cite{xu2022bits}.
Synthesized pedestrian behavior can also be controlled by strategically choosing a starting location~\cite{priisalu2022generating}.
STRIVE~\cite{rempe2022strive} showed that a VAE trajectory model can be controlled through test-time optimization in the learned latent space.
Reinforcement learning (RL) agents can be controlled in crowd simulations by incorporating tasks into reward functions for training~\cite{lee2018crowd}.
By varying the weights of different rewards, the characters can be controlled to exhibit one of several behaviors at test time~\cite{panayiotou2022ccp}.
Our method, \name, trains %
to mimic trajectories from data and is agnostic to any task: all controls are defined at test time, allowing flexibility to new controls after training. Instead of lengthy test-time optimization, we use guidance for control.

\paragraph{Controllable Character Animation}
Full-body pedestrian animation typically involves a high-level task (\eg trajectory following, obstacle avoidance) and low-level body control. 
Some methods solve both with a single network that implicitly uses high-level planning and low-level animation. GAMMA~\cite{zhang2022wanderings} trains a kinematic model to go to waypoints, while PFNN~\cite{holden2017phase} follows gamepad inputs. Physics-based humanoid controllers such as AMP~\cite{peng2021amp} train different models for each task, limiting their general applicability. 

Two-stage methods split the task into separate high-level planning and low-level character control, where task information is only used by the planner. 
Planning can be done with traditional A*~\cite{hassan2021stochastic}, using learned trajectory prediction~\cite{cao2020long}, searching in a pre-trained latent space~\cite{ling2020character,won2022physics,peng2022ase,rempe2021humor}, or using hierarchical RL \cite{Haworth2020DeepIO, peng2017deeploco, peng2022ase, priisalu2020semantic,won2022physics}. DeepLoco \cite{peng2017deeploco}, Haworth \etal \cite{Haworth2020DeepIO}, and ASE \cite{peng2022ase} utilize hierarchical RL to achieve impressive dynamic control for various tasks. They require lengthy training for both low-level and high-level controllers and often jointly train as a final step. They must also train different planners for different tasks.

Our approach follows the two-stage paradigm, with the distinction that both our high-level (\name) and low-level (\animname) models consume task information for pedestrian navigation: through test-time guidance and map-conditioned path following, respectively.  %
\name and \animname are unaware of each other at training time, yet can be tightly integrated in a closed loop: trace-pace-retrace.

\paragraph{Diffusion Models and Guidance}
Diffusion models have shown success in generating images~\cite{ho2020denoising,nichol2021improved,vahdat2021score}, videos~\cite{ho2022imagen}, and point clouds~\cite{zeng2022lion}.
Guidance has been used for test-time control in several ways: classifier~\cite{dhariwal2021diffusion} and classifier-free~\cite{ho2022classifier} guidance reinforce input conditioning, while reconstruction guidance~\cite{ho2022video} has been used for coherent video generation.
Gu \etal ~\cite{gu2022stochastic} adapt the diffusion framework for short-term pedestrian trajectory forecasting conditioned on past trajectories.
Diffuser~\cite{janner2022diffuser} generates trajectories for planning and control in robotics applications with test-time guidance.
Closest to ours is the concurrent work of CTG~\cite{zhong2022ctg}, which builds on Diffuser to develop a controllable vehicle traffic model, focusing on following formalized traffic rules like speed limits. 
Our method contains several key differences: we encode map conditioning into an expressive feature grid queried in denoising, we use classifier-free sampling to enable multi-dataset training and test-time flexibility, we re-formulate guidance to operate on clean model outputs, and we link with a low-level animation model using value function guidance. 

\vspace{-2mm}
\section{Method}
\label{sec:method-diffusion}
To model high-level pedestrian behavior, we first introduce the controllable trajectory diffusion model (\name).
In \cref{sec:method-amp}, we detail our low-level physics-based pedestrian controller, \animname, and in \cref{sec:method-system} how they can be combined into an end-to-end animation system.

\vspace{-0.5mm}
\subsection{Controllable Trajectory Diffusion}
\vspace{-0.5mm}

\paragraph{Problem Setting}
Our goal is to learn high-level pedestrian behavior in a way that can be \textit{controlled} at test time.
For pedestrian animation, we focus on two types of control: (1) user specification, \eg, goal waypoints, social distance, and social groups, and (2) physical plausibility, \eg, avoiding collisions with obstacles or between pedestrians.

We formulate synthesizing pedestrian behavior as an agent-centric trajectory forecasting problem.
At each time step, the model outputs a future trajectory plan for a target \textit{ego} agent conditioned on that agent's past, the past trajectories of all neighboring agents, and the semantic map context.
Formally, at timestep $t$ we want the future state trajectory $\btau_s = [ \bes_{t+1} \quad \bes_{t+2} \quad \dots \quad \bes_{t+T_f}]$ over the next $T_f$ steps where the state $\bes = [x \quad y \quad \theta \quad v]^T$ includes the 2D position $(x,y)$, heading angle $\theta$, and speed $v$.
We assume this state trajectory is actually the result of a sequence of actions~\cite{zhong2022ctg} defined as $\btau_a = [ \ba_{t+1} \quad \ba_{t+2} \quad \dots \quad \ba_{t+T_f}]$ where each action $\ba = [\dot{v} \quad \dot{\theta}]^T$ contains the acceleration $\dot{v}$ and yaw rate $\dot{\theta}$.
The state trajectory can be recovered from the initial state and action trajectory as $\btau_s = f(\bes_t, \btau_a)$ using a given dynamics model $f$.
The full state-action trajectory is then denoted as $\btau = [\btau_s; \btau_a]$. 
To predict the future trajectory, the model receives as input the past state trajectory of the ego pedestrian $\bx^\text{ego} = [ \bes_{t-T_p} \quad \dots \quad \bes_{t}]$ along with the past trajectories of $N$ neighboring pedestrians $X^\text{neigh} = \{\bx^\text{i} \}_{i=1}^{N}$.
It also gets a crop of the rasterized semantic map $\map \in \reals^{H \times W \times C}$ in the local frame of the ego pedestrian at time $t$.
These inputs are summarized as the \textit{conditioning} context $C = \{ \bx^\text{ego}, X^\text{neigh}, \map \}$.

Our key idea is to train a diffusion model to conditionally generate trajectories, which can be \textit{guided} at test time to enable controllability.
For simplicity, the following formulation uses the full trajectory notation $\btau$, but in practice, the state trajectory is always a result of actions, \ie, diffusion/denoising are on $\btau_a$ which determines the states through $f$. %
Next, we summarize our diffusion framework, leaving the details to the supplementary material.

\begin{figure*}
\vspace{-1.5mm}
\begin{center}
\includegraphics[width=0.95\textwidth]{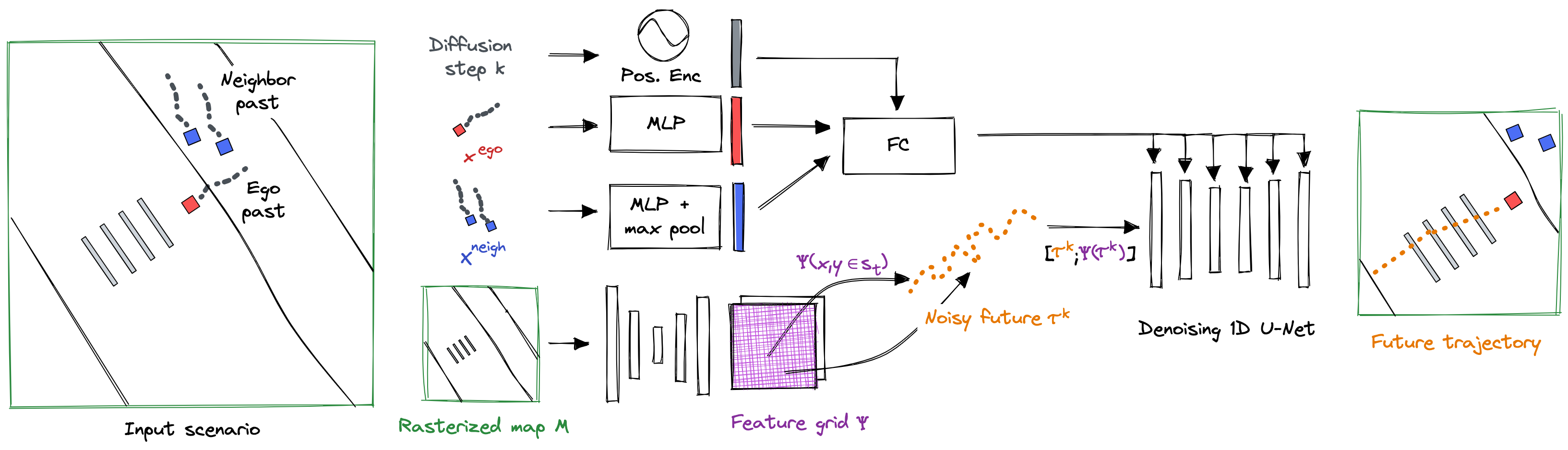}
\end{center}
\vspace{-5mm}
   \caption{Trajectory diffusion model (\name). Future trajectory denoising is conditioned on past and neighbor motion by adding processed features to intermediate U-Net features. Map conditioning is provided through a feature grid queried along the noisy input trajectory.
   }
\vspace{-3.5mm}
\label{fig:arch}
\end{figure*}

\vspace{-3mm}
\subsubsection{Trajectory Diffusion Model}
\vspace{-1mm}
We build on Diffuser~\cite{janner2022diffuser} and generate trajectories through iterative denoising, which is learned as the reverse of a pre-defined \textit{diffusion} process~\cite{ho2020denoising,sohl2015deep}.
Starting from a clean future trajectory $\btau^0 \sim q(\btau^0)$ sampled from the data distribution, the forward noising process produces a sequence of progressively noisier trajectories $(\btau^1, \dots, \btau^k, \dots, \btau^K)$ by adding Gaussian noise at each process step $k$:
\vspace{-3mm}
\begin{equation}
\small
\begin{aligned}
    q(\btau^{1:K} \mid \btau^0) &:= \prod_{k=1}^K q(\btau^k \mid \btau^{k-1}) \\
    q(\btau^k \mid \btau^{k-1})  &:= \normal (\btau^k ; \sqrt{1-\beta_k} \btau^{k-1}, \beta_k \mathbf{I})
\end{aligned}
\end{equation}
where $\beta_k$ is the variance at each step of a fixed schedule, and with a large enough $K$ we get $q(\btau^K) \approx \normal(\btau^K; \mathbf{0}, \mathbf{I})$.
\name learns the reverse of this process so that the sampled noise can be \textit{denoised} into plausible trajectories.
Each step of this reverse process is conditioned on $C$:
\begin{equation}
\label{eqn:prior}
    p_\phi(\btau^{k-1} \mid \btau^k, C) := \normal (\btau^{k-1} ; \bmu_\phi(\btau^k, k, C), \boldsymbol{\Sigma}_k)
\end{equation}
where $\phi$ are model parameters and $\boldsymbol{\Sigma}_k$ is from a fixed schedule.
\name learns to parameterize the mean of the Gaussian distribution at each step of the denoising process.

\paragraph{Training and Classifier-Free Sampling}
Importantly for guidance, the network does \emph{not} directly output $\bmu$. Instead, at every step it learns to predict the final clean trajectory $\btau^0$, which is then used to compute $\bmu$~\cite{nichol2021improved}.
Training supervises this network output $\hat{\btau}^0$ with ground truth future trajectories (\ie denoising score matching~\cite{vincent2011connection,song2019generative,ho2020denoising}):
\begin{equation}
    L = \mathbb{E}_{\beps, k, \btau^0, C} \left[ || \btau^0 - \hat{\btau}^0 ||^2 \right]
\end{equation}
where $\btau^0$ and $C$ are sampled from the training dataset, $k \sim \mathcal{U}\{1,2,\dots,K\}$ is the step index, and $\beps \sim \normal (\mathbf{0}, \mathbf{I})$ is used to corrupt $\btau^0$ to give the noisy input trajectory $\btau^k$.

Our training procedure allows the use of \textit{classifier-free sampling}\footnote{we refer to it as ``sampling'' instead of the common term ``guidance''~\cite{ho2022classifier} to avoid confusion with the guidance introduced in \cref{sec:guidance}} at test time, which has been shown to improve compliance to conditioning in diffusion models~\cite{ho2022classifier}.
We simultaneously train both a conditional model $\bmu_\phi(\btau^k, k, C)$ and unconditional model $\bmu_\phi(\btau^k, k)$ by randomly dropping out conditioning during training.
At test time, predictions from both models are combined with weight $w$ as:
\begin{equation}
    \tilde{\beps}_\phi = \beps_\phi(\btau^k, k, C) + w \left( \beps_\phi(\btau^k, k, C) - \beps_\phi(\btau^k, k) \right)
\end{equation}
where $\beps_\phi$ is the model's prediction of how much noise was added to the clean trajectory to produce the input $\btau^k$; it is straightforward to compute from $\bmu_\phi$~\cite{nichol2021improved}.

Note that $w$$>$$0$ and $w$$<$$0$ increase and decrease the effect of conditioning, respectively, while $w$$=$$0$ and $w$$=$$-1$ result in the purely conditional or unconditional model, respectively.
This flexibility allows a user to trade off respecting conditioning with trajectory diversity, which benefits controllability (see \cref{sec:results-realworld}).
This approach also enables training on multiple distinct datasets with varying annotations: conditioning is already being dropped out randomly, so it is easy to use mixed data with subsets of the full conditioning.
Since there are pedestrian datasets with diverse motions but no semantic maps~\cite{pellegrini2009you,lerner2007crowds}, and others with limited motions but detailed maps~\cite{caesar2020nuscenes}, we find mixed training is beneficial to boost diversity and controllability (see \cref{sec:results-realworld}).

\paragraph{Architecture}
As shown in \cref{fig:arch}, \name uses a U-Net similar to~\cite{janner2022diffuser} that has proven effective for trajectories. %
The input trajectory $\btau_k$ at step $k$ is processed by a sequence of 1D temporal convolutional blocks that progressively down and upsample the sequence in time, leveraging skip connections.
A key challenge is how to condition the U-Net on $C$ to predict trajectories that comply with the map and other pedestrians.
To incorporate step $k$, ego past $\bx^\text{ego}$, and neighbor past $X^\text{neigh}$, we use a common approach~\cite{janner2022diffuser,ho2022video} that extracts a single conditioning feature and adds it to the intermediate trajectory features within each convolutional block.
For the map $\map$, we encode with a 2D convolutional network into a feature grid, where each pixel contains a high-dimensional feature.
At step $k$ of denoising, each 2D position $(x,y) \in \btau^k$ is queried by interpolating into the grid to give a feature trajectory, which is concatenated to $\btau^k$ and becomes the U-Net input.
Intuitively, this allows learning a localized representation that can benefit subtle map interactions such as obstacle avoidance.

\vspace{-2mm}
\subsubsection{Controllability through Clean Guidance}
\label{sec:guidance}
After training \name to generate realistic trajectories, controllability is implemented through test-time \emph{guidance}.
Intuitively, guidance nudges the sampled trajectory at each step of denoising towards a desired outcome.
Let $\guide (\btau)$ be a guidance loss function measuring how much a trajectory $\btau$ violates a user objective.
This may be learned~\cite{janner2022diffuser} or an analytical differentiable function~\cite{zhong2022ctg}.
Guidance uses the gradient of $\guide$ to perturb the predicted mean from the model at each denoising step such that the right side of \cref{eqn:prior} becomes $\normal (\btau^{k-1} ; \tilde{\bmu}_\phi(\btau^k, k, C), \boldsymbol{\Sigma}_k)$ where $\tilde{\bmu}$ is the perturbed (guided) mean. 
Prior work~\cite{janner2022diffuser,zhong2022ctg} directly perturbs the \emph{noisy} network-predicted mean with
\begin{equation}
\label{eqn:noisy-guide}
    \tilde{\bmu} = \bmu - \alpha \boldsymbol{\Sigma}_k \nabla_{\bmu} \guide (\bmu)
\end{equation}
where $\alpha$ determines the guidance strength. Note that \cref{eqn:noisy-guide} evaluates $\guide$ at the noisy mean, so learned loss functions must be trained at varying noise levels and analytic loss functions may suffer from numerical issues. %

To avoid this, we build upon ``reconstruction guidance'', which operates on the \emph{clean} model prediction $\hat{\btau}^0$~\cite{ho2022video}.
We extend the guidance formulation introduced in \cite{ho2022video} for temporal video upsampling to work with arbitrary loss functions. %
At each denoising step with input $\btau^k$, we first perturb the clean trajectory predicted from the network $\hat{\btau}^0$ with
\begin{equation}
\label{eqn:clean-guide}
    {\tilde{\btau}^0} = \hat{\btau}^0 - \alpha \boldsymbol{\Sigma}_k \nabla_{\btau^k} \guide (\hat{\btau}^0),
\end{equation}
then compute $\tilde{\bmu}$ in the same way as we would in \cref{eqn:prior}, \ie, as if $\tilde{\btau}^0$ were the output of the network.
Note that the gradient is evaluated wrt the noisy input trajectory $\btau^k$ rather than the clean $\hat{\btau}^0$, requiring backpropagation through the denoising model.
We formulate several analytical guidance objectives like waypoint reaching, obstacle avoidance, collision avoidance, and social groups (see \cref{sec:results-orca}, \ref{sec:results-realworld}). A learned RL value function can also be used (\cref{sec:results-anim}). 

\subsection{Physics-Based Pedestrian Animation}
\label{sec:method-amp}

To enable full-body pedestrian simulation, we design the Pedestrian Animation ControllER (PACER) to execute the 2D trajectories generated by \name in a physics simulator.

\begin{figure}
\vspace{-0mm}
    \centering
    \includegraphics[width=\linewidth]{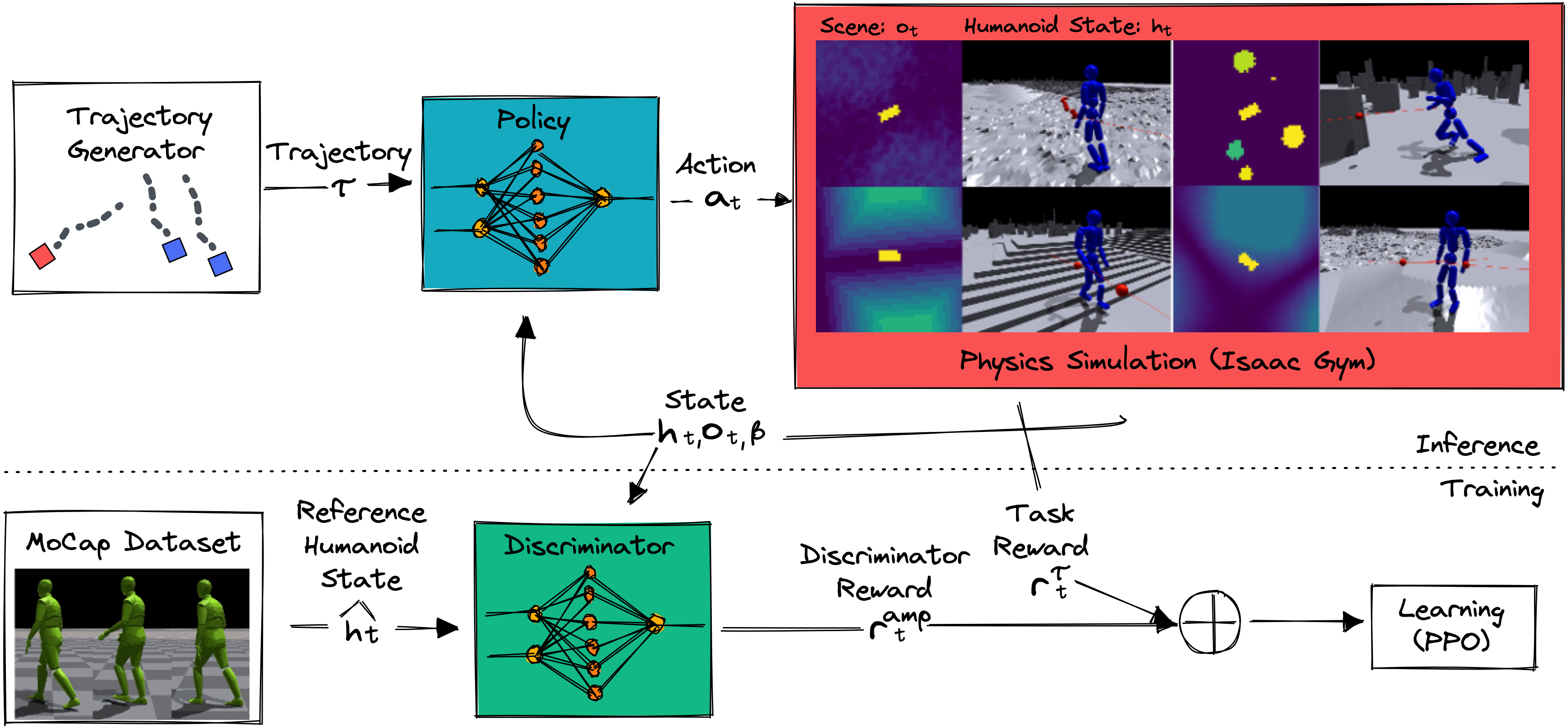}
    \vspace{-4mm}
    \caption{
    \small
    Pipeline: Pedestrian Animation Controller (PACER).}
\vspace{-5mm}
    \label{fig: pacer_train}
\end{figure}

\paragraph{Background: Goal-Conditioned RL} 
Our framework (\cref{fig: pacer_train}) follows the general goal-conditioned reinforcement learning framework, where a goal-conditioned policy $\pi_{\text{PACER}}$ is trained to follow 2D target trajectories specified by $\btau_s$. The task is formulated as a Markov Decision Process (MDP) defined by a tuple ${\mathcal M}=\langle \mathcal{S}, \mathcal{A}, \mathcal{T}, R, \gamma\rangle$ of states, actions, transition dynamics, reward function, and discount factor. The state $\mathcal{S}$, transition dynamics $\mathcal{T}$, and reward $R$ are calculated by the environment based on the current simulation and goal, while the action $\mathcal{A}$ is computed by the policy $\pi_{\text{PACER}}$. The policy's objective is to maximize the discounted return $\mathbb{E}\left[\sum_{t=1}^{T} \gamma^{t-1} r_{t}\right]$ where $r_t$ is the reward per timestep. 
We utilize Proximal Policy Optimization (PPO)~\cite{Schulman2017PROXimalPO} to find the optimal control policy $\pi_{\text{PACER}}$.

\paragraph{Terrain, Social, and Body Awareness}
To create a controller that can simulate crowds in realistic 3D scenes (\eg scans, neural reconstructions, or artist-created meshes (\cref{fig:teaser})), our humanoid must be terrain aware, socially aware of other agents, and support diverse body types. 
We use a humanoid model that conforms to the kinematic structure of SMPL~\cite{Loper2015SMPLAS}, and is automatically generated using a procedure similar to~\cite{yuan2021simpoe, Luo2021DynamicsRegulatedKP, Luo2022EmbodiedSH}.
Our control policy $\pi_{\text{PACER}}(\bs a_t | \bs h_t, \bs o_t, \bs{\beta}, \btau_s)$ is conditioned on the state of the simulated character $\bs h_t$, environmental features $\bs o_t$, body type $\bs \beta$, and goal trajectory $\bs \tau_s$. The environment input is a rasterized local height and velocity map of size $\bs o_t \in \mathbb{R}^{64 \times 64 \times 3}$, which gives agents crucial information about their surroundings. To allow for social awareness, nearby humanoids are represented as a cuboid and rendered on the global height map. In this way, each humanoid views other people as dynamic obstacles to avoid. Obstacle and interpersonal avoidance are learned by using obstacle collision as a termination condition. By conditioning and training with different body parameters $\bs{\beta}$ our policy learns to adapt to characters with diverse morphologies. 

\paragraph{Realistic Motion through Adversarial Learning}
To learn the optimal control policy $\pi_{\text{PACER}}$ that (1) follows a 2D trajectory closely and (2) creates realistic pedestrian motions, we follow Adversarial Motion Prior (AMP)~\cite{peng2021amp}. AMP uses a motion discriminator to encourage the policy to generate motions that are similar to the movement patterns contained in a dataset of motion clips recorded by human actors. The discriminator $D(\bs h_{t-10:t}, \bs a_t)$ is then used to specify a motion style reward $r^{\text{amp}}_t$ for training the policy.  The style reward is combined with a trajectory following reward $r^{\tau}_t$ and an energy penalty $r^{\text{energy}}_t$ \cite{Fu2022DeepWC} to produce the total reward $r_t = r^{\text{amp}}_t + r^{\tau}_t +  r^{\text{energy}}_t$. To mitigate artifacts arising from asymmetric gaits, such as limping, we utilize the motion-symmetry loss proposed by \cite{Yu2018LearningSA}:
\vspace{-1mm}
\begin{equation}
\small
\begin{aligned}
    L_{\text{sym}}(\theta)= &\|\pi_\text{PACER}(\bs h_t, \bs o_t, \bs \beta, \btau_s) - \\
    &\Phi_a(\pi_\text{PACER}(\Phi_s(\bs h_t, \bs o_t, \bs \beta, \btau_s)))\|^2,  
\end{aligned}
\end{equation}
where $\Phi_s$ and $\Phi_a$ mirror the state and action along the character's sagittal plane. This loss encourages the policy to produce more symmetric motions, leading to natural gaits. During training, random terrains are generated following the procedure used in \cite{rudin2021learning}. We create stairs, slopes, uneven terrains, and obstacles consisting of random polygons. 
Character morphology is also randomized by sampling a gender and body type from the AMASS dataset \cite{Mahmood2019AMASSAO}. The policy and discriminator are then conditioned on the SMPL gender and body shape $\bs\beta$ parameters. 
More details are available in the supplementary material.

\subsection{Controllable Pedestrian Animation System}
\label{sec:method-system}
The high-level trajectory planning from \name is combined with the low-level character control from \animname to create an end-to-end pedestrian animation system.
The two components are trained independently, but at runtime they operate in a closed feedback loop: \animname follows planned trajectories for $2s$ before \name re-planning, taking past character motion from PACER as input.
By combining terrain and social awareness of \animname with collision avoidance guidance, both high and low-level systems are task-aware and work in tandem to prevent collisions and falls.

\paragraph{Value Function as Guidance}
To enable tighter two-way coupling between \name and \animname, in \cref{sec:results-anim} we explore using the value function learned during RL training of \animname to guide trajectory diffusion.
The value function predicts expected future rewards and is aware of body pose and surrounding terrain and agents. Using the value function to guide denoising encourages \name to produce trajectories that are easier to follow and better suited to the current terrain (which \name is unaware of otherwise).
Unlike Diffuser~\cite{janner2022diffuser}, which requires training a reward function with samples from the diffusion model at varying noise levels, our guidance (\cref{eqn:clean-guide}) operates on clean trajectories so we can use the value function directly from RL training.
\vspace{-2mm}
\section{Experiments}
\label{sec:results}
We first demonstrate the controllability of \name when trained on synthetic (\cref{sec:results-orca}) and real-world (\cref{sec:results-realworld}) pedestrian data.
\cref{sec:results-anim} evaluates our full animation system on several tasks and terrains.
\textbf{Video results} are provided in the supplementary material.

\begin{figure}[t]
\begin{center}
\includegraphics[width=0.95\linewidth]{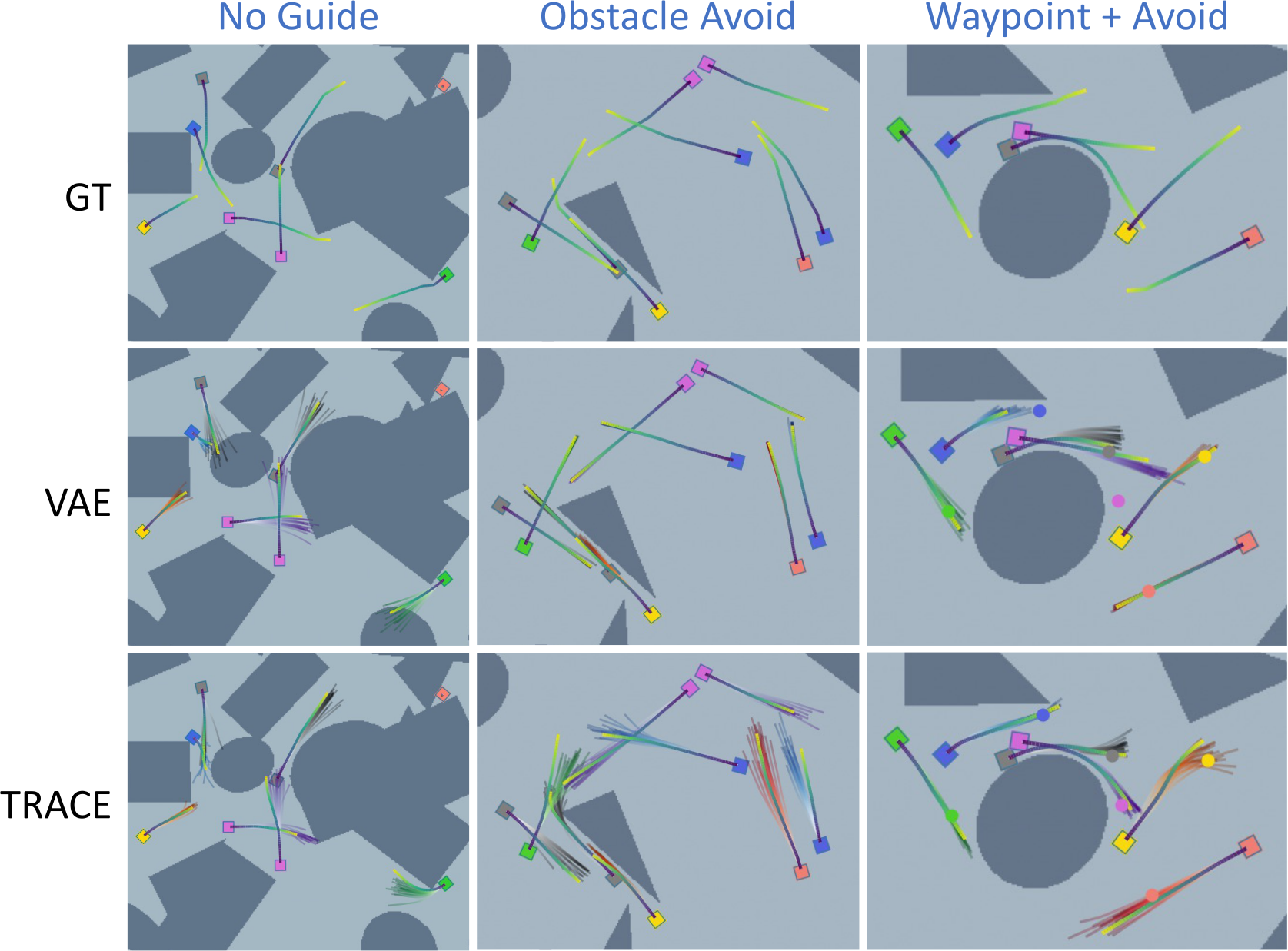}
\vspace{-6mm}
\end{center}
  \caption{Guidance results on \textit{ORCA-Maps}. For VAE and \name, 20 samples are visualized for each pedestrian (the boxes) along with the final trajectory chosen via filtering which is bolded.
  }
\vspace{-4.5mm}
\label{fig:orca-qual}
\end{figure}

\paragraph{Implementation Details}
\name is trained to predict $5s$ of future motion from $3s$ of past motion (both at $10$Hz), and uses $K{=}100$ diffusion steps.
During training, map and neighbor conditioning inputs are independently dropped with $10\%$ probability.
At test time, we sample (and guide) multiple future trajectories for each pedestrian in a scene and choose one with the lowest guidance loss, which we refer to as \textit{filtering}. PACER operates at $30$Hz; we randomly sample terrain, body type, and procedural 2D trajectories during training and use a dataset of locomotion sequences from AMASS \cite{Mahmood2019AMASSAO}. All physics simulations are performed using NVIDIA's Isaac Gym simulator \cite{makoviychuk2021isaac}.

\paragraph{Datasets}
The \textit{ORCA} dataset (\cref{sec:results-orca}) contains synthetic trajectory data from $10s$ scenes generated using the ORCA crowd simulator~\cite{berg2011reciprocal}. Up to 20 agents are placed in a $15m{\times}15m$ environment with $\leq$20 static primitive obstacles. Agent placement and goal velocity, along with obstacle placement and scale, are randomized per scene. The dataset contains two distinct subsets: \textit{ORCA-Maps} has many obstacles but few agents, while \textit{ORCA-Interact} has no obstacles (\ie no map annotations) but many agents.

For real-world data (\cref{sec:results-realworld}), we use ETH/UCY and nuScenes.
ETH/UCY~\cite{pellegrini2009you,lerner2007crowds} is a common trajectory forecasting benchmark that contains scenes with dense crowds and interesting pedestrian dynamics but does not have semantic maps.
nuScenes~\cite{caesar2020nuscenes} contains $20s$ driving scenes in common street settings. We convert the pedestrian bounding-box annotations to 2D trajectories and use them for training and evaluation. %
Detailed semantic maps are also annotated with layers for roads, crosswalks, and sidewalks.

\paragraph{Metrics}
We care about trajectory plausibility and meeting user controls.
Controllability is evaluated with a \textit{Guidance Error} that depends on the task: \eg, for avoidance objectives this is collision rate, while the waypoint error measures the minimum distance from the trajectory. 
\textit{Obstacle and Agent Collision Rates} measure the frequency of collisions.
\textit{Realism} is measured at the dataset or trajectory level by (1) computing the Earth Mover's Distance (EMD) between the generated and ground truth test-set histograms of trajectory statistics (\eg velocity, longitudinal/lateral acceleration)~\cite{xu2022bits}, or (2) measuring the mean accelerations of each trajectory, assuming pedestrians generally move smoothly.

\begin{table}[t]
\begin{center}
\scalebox{0.6}{
\begin{tabular}{ll | c | c c | c c c}
\toprule
 & & \textbf{Guidance} & \multicolumn{2}{c}{\textbf{Collision Rate}} & \multicolumn{3}{|c}{\textbf{Realism} \small{(EMD)}} \\
\textbf{Guide} & \textbf{Method} & Error &  Obstacle & Agent & Vel & Lon Acc  & Lat Acc \\
\midrule
None & VAE~\cite{rempe2022strive} & -- & 0.076 & \textbf{0.118} & 0.038 & 0.039 & 0.040 \\
     & \name & -- & \textbf{0.050} & 0.132 & \textbf{0.029} & \textbf{0.008} & \textbf{0.009} \\
\midrule
Obstacle & VAE~\cite{rempe2022strive} & 0.018 & 0.018 & \textbf{0.116} & 0.040 & 0.036 & 0.039 \\
Avoid & \name-Filter & 0.018 & 0.018 & 0.123 & \textbf{0.019} & \textbf{0.011} & \textbf{0.015} \\
         & \name-Noisy & 0.015 & 0.015 & 0.125 & 0.021 & 0.012 & 0.017 \\
         & \name & \textbf{0.014} & \textbf{0.014} & 0.124 & 0.020 & \textbf{0.011} & 0.017 \\
\midrule
Agent & VAE~\cite{rempe2022strive} & 0.010 & 0.075 & 0.010 & 0.041 & 0.038 & 0.039 \\
Avoid & \name-Filter & 0.049 & \textbf{0.050} & 0.049 & 0.031 & 0.012 & 0.013 \\
         & \name-Noisy & \textbf{0.000} & 0.056 & \textbf{0.000} & 0.035 & 0.013 & \textbf{0.012} \\
         & \name & \textbf{0.000} & 0.058 & \textbf{0.000} & \textbf{0.025} & \textbf{0.010} & \textbf{0.012} \\
\midrule
Waypoint & VAE~\cite{rempe2022strive} & \textbf{0.078} & 0.051 & \textbf{0.092} & 0.070 & 0.031 & 0.033 \\
         & \name-Filter & 0.333 & \textbf{0.046} & 0.112 & \textbf{0.044} & \textbf{0.013} & \textbf{0.013} \\
         & \name-Noisy & 0.129 & 0.052 & 0.110 & 0.067 & 0.038 & 0.033 \\
         & \name & 0.105 & 0.048 & 0.093 & 0.057 & \textbf{0.013} & 0.014 \\
\midrule
Waypoint & VAE~\cite{rempe2022strive} & \textbf{0.207} & \textbf{0.021} & 0.015 & 0.053 & 0.032 & 0.032 \\
\& Obs Avoid & \name-Filter & 0.527 & 0.023 & 0.096 & \textbf{0.025} & 0.014 & 0.016 \\
\& Agt Avoid & \name-Noisy & 0.236 & 0.022 & 0.017 & 0.057 & 0.025 & 0.022 \\
            & \name & 0.211 & \textbf{0.021} & \textbf{0.009} & 0.036 & \textbf{0.007} & \textbf{0.009} \\
\bottomrule
\end{tabular}}
\end{center}
\vspace{-6mm}
\caption{Guidance evaluation on \textit{ORCA-Maps} dataset. \name using full diffusion guidance improves upon VAE latent optimization and selective sampling (\textit{\name-Filter}) in terms of meeting objectives, while maintaining strong realism.  
}
\label{tab:orca-compare}
\vspace{-3mm}
\end{table}

\begin{table}
\begin{center}
\scalebox{0.7}{
\begin{tabular}{lll c c c c}
\toprule
 &  &  & & \textbf{Guidance} & \multicolumn{2}{c}{\textbf{Realism} \small{(Mean)}} \\
\textbf{Guide} & \textbf{Method} & \textbf{Train Data} & $w$ & Error & Lon Acc & Lat Acc \\
\midrule
Waypoint & VAE~\cite{rempe2022strive} & Mixed & --  & \textbf{0.340} & 0.193 & 0.172\\
         & \name & nuScenes & -0.5  & 0.421 & 0.177 & 0.168 \\
         & & Mixed & 0.0 &   0.551 & 0.159 & 0.145 \\
         & & Mixed &  -0.5 &  0.366 & \textbf{0.140} & \textbf{0.132} \\
\midrule
Waypoint & VAE~\cite{rempe2022strive} & Mixed & --  & 0.962 & 0.443 & 0.441\\
perturbed & \name & nuScenes & -0.5  & 0.977 & 0.239 & 0.238 \\
            & & Mixed & 0.0 &   1.129 & 0.233 & 0.218 \\
            & & Mixed &  -0.5 & \textbf{0.802} & \textbf{0.212} & \textbf{0.204} \\
\midrule
Social   & VAE~\cite{rempe2022strive} & Mixed & --  & 0.297 & 0.109 & 0.104\\
groups   & \name & nuScenes & -0.5  & 0.287 & 0.155 & 0.158 \\
        & & Mixed & 0.0 &  \textbf{0.244} & 0.110 & 0.101 \\
         & & Mixed &  -0.5 &  0.245 & \textbf{0.094} & \textbf{0.087} \\
\bottomrule
\end{tabular}}
\end{center}
\vspace{-6mm}
\caption{Guidance evaluation on nuScenes. Training on mixed data and using $w$$<$$0$ for classifier-free sampling are important to achieve controllability for out-of-distribution objectives.
}
\label{tab:real-compare}
\vspace{-6mm}
\end{table}
\begin{figure}
\begin{center}
\includegraphics[width=0.95\linewidth]{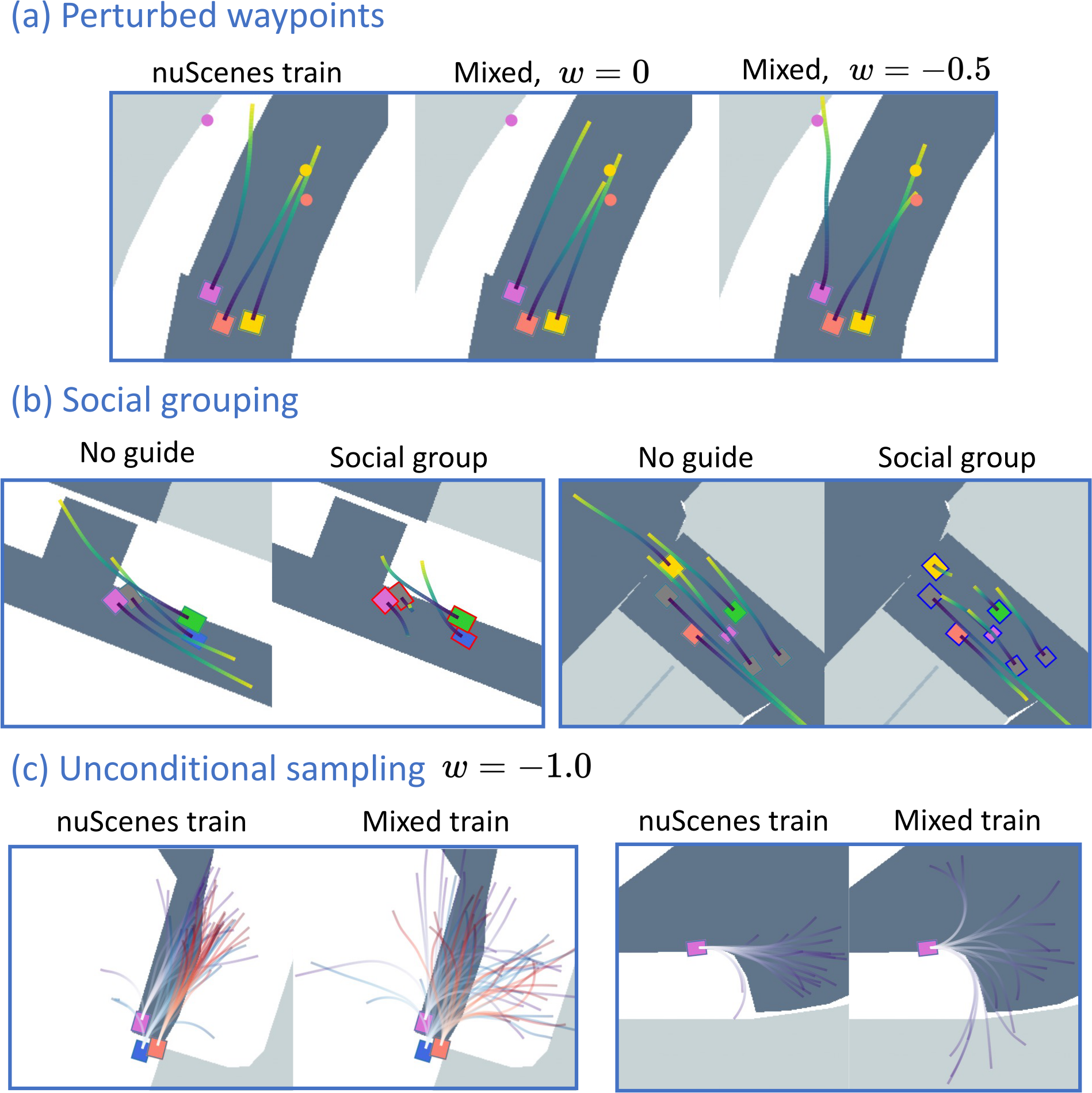}
\vspace{-5mm}
\end{center}
  \caption{nuScenes results demonstrating flexibility of \name. (a) Using mixed training and 
$w$$=$$-0.5$ is best for noisy waypoints. (b) Social group guidance encourages sets of pedestrians to stay close. (c) Mixed training (ETH/UCY+nuScenes) learns a more diverse distribution as demonstrated by unconditional sampling.}
\vspace{-4mm}
\label{fig:real-qual}
\end{figure}

\begin{figure*}[t]
\begin{center}
\includegraphics[width=1.0\linewidth]{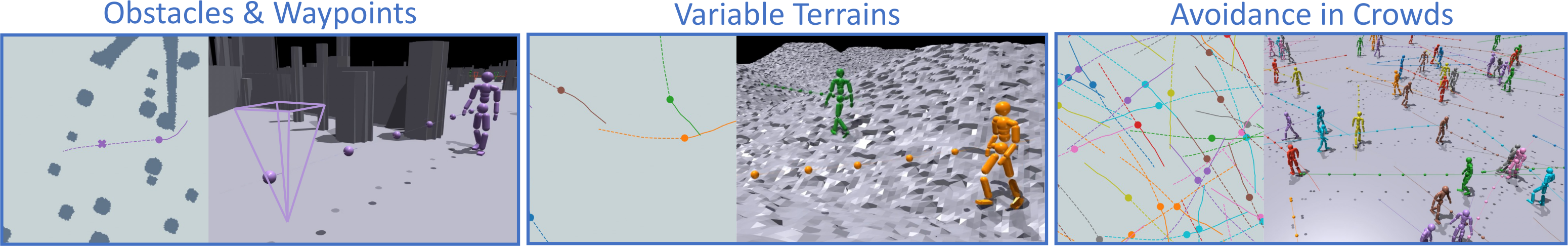}
\vspace{-10mm}
\end{center}
  \caption{Our animation system enables avoiding obstacles, meeting goals, traversing variable terrains, and large crowds. 
  }
\vspace{-5mm}
\label{fig:anim-qual}
\end{figure*}

\subsection{Augmenting Crowd Simulation}
\label{sec:results-orca}
We first evaluate \name trained on \textit{ORCA-Maps} and \textit{ORCA-Interact}. These provide a clean test bed for comparisons since there is a clear definition of correct pedestrian behavior -- no obstacle or agent collisions are present in the data.
All methods operate in an open loop by predicting a single $5s$ future for each pedestrian. This way, compounding errors inherent to closed-loop operation are not a factor.

Results for single and multi-objective guidance on the \textit{ORCA-Maps} test set are shown in \cref{tab:orca-compare}. 
\name is compared to a \textit{VAE} baseline~\cite{rempe2022strive} adapted to our setup, which achieves controllability through test-time latent optimization. This is a strong baseline that generally works well, but requires expensive optimization at test time.
We also compare to two ablations: \textit{\name-Filter} samples from the diffusion model \emph{without guidance} and chooses the best sample according to the guidance loss (similar to \cite{xu2022bits}), while \textit{\name-Noisy} uses the guidance formulated in \cref{eqn:noisy-guide} from prior works~\cite{janner2022diffuser,zhong2022ctg}.
Models are trained on the combined dataset of \textit{ORCA-Maps} (with map annotations) and \textit{ORCA-Interact} (no map annotations).
The guidance losses are: 
\textbf{None} samples randomly with no guidance; \textbf{Obstacle avoid} discourages collisions between map obstacles and pedestrian bounding boxes; \textbf{Agent avoid} discourages collisions between pedestrians by denoising all their futures in a scene jointly; \textbf{Waypoint} encourages a trajectory to pass through a goal location at any point in the planning horizon. For this experiment, the waypoint is set as the position of each pedestrian at $4s$ into the future in the ground truth data.
These are \emph{in-distribution} objectives, since they reinforce behavior already observed in the ground truth data. 

In \cref{tab:orca-compare}, \name successfully achieves all objectives through the proposed guidance. 
It is competitive or better than the VAE optimization in terms of guidance, while maintaining velocity and acceleration distributions closer to ground truth as indicated by \textit{Realism}.
\cref{fig:orca-qual} shows that random samples from the VAE contain collisions, and using latent optimization for controllability gives similar local minima across samples thereby limiting diversity compared to \name. 
Finally, using our proposed clean guidance (\cref{eqn:clean-guide}) instead of the noisy version produces consistently better results in guidance and realism.

\subsection{Real-world Data Evaluation}
\label{sec:results-realworld}
We next evaluate controllability when trained on real-world data, and focus on \emph{out-of-distribution} (OOD) guidance objectives to emphasize the flexibility of our approach.
In this experiment, methods operate in a \emph{closed loop}: pedestrians are rolled out for $10s$ and re-plan at 1Hz. 
Results on a held out nuScenes split are shown in \cref{tab:real-compare}. We compare \name trained on mixed data (ETH/UCY+nuScenes), after training on nuScenes only, and using two different classifier-free sampling weights $w$.
Along with in-distribution \textbf{Waypoint} (now at $9s$ into the future), two additional objectives are evaluated: \textbf{Waypoint perturbed} uses a noisily perturbed ground truth future position (at $9s$), requiring pedestrians to go off sidewalks or into streets to reach the goal; \textbf{Social groups} specifies groups of agents to stay close and travel together. Groups are set heuristically based on spatial proximity and velocity at initialization.

In \cref{tab:real-compare}, we observe that OOD flexibility requires (1) training on mixed data, and (2) classifier-free sampling. 
Since nuScenes data is less diverse (people tend to follow the sidewalk), \name trained on just nuScenes struggles to hit perturbed waypoints.
Though the VAE is trained on mixed data, it struggles to produce diverse dynamics on the nuScenes maps to achieve OOD objectives, even though it uses 200 optimization steps (2$\times$ more than the diffusion steps $K{=}100$ in \name).
\name reaches OOD objectives using classifier-free sampling with $w$$=$$-0.5$ to downweight the conditioning of the semantic map and leverage diverse trajectories learned from ETH/UCY.
The flexibility of \name is further highlighted in \cref{fig:real-qual}.

\subsection{Controllable Pedestrian Animation}
\label{sec:results-anim}
Finally, we demonstrate our full controllable pedestrian animation system.
\name is trained on \textit{ORCA} and used as a planner for the pre-trained \animname without any fine-tuning.
We evaluate the animations by: \textit{Fail Rate} measures the fraction of agents that fall down or collide with an obstacle or other agent, \textit{Trajectory Following Error} measures the average deviation of the character from \name's plan, and \textit{Discriminator Reward} is the mean reward returned by the adversarial motion prior used to train \animname, which measures how human-like a generated motion appears.
 
\begin{table}
\begin{center}
\scalebox{0.8}{
\begin{tabular}{ll c c c}
\toprule
& & \textbf{Fail} & \textbf{Traj Follow} & \textbf{Discrim} \\
\textbf{Terrain} & \textbf{Guide} & Rate & Error & Reward \\
\midrule
Random & Procedural & 0.133 & 0.680 & 1.950 \\
       & None & \textbf{0.093} & \textbf{0.104} & 1.887 \\
       & Waypoint & 0.107 & 0.111 & \textbf{2.113}\\
\midrule
Obstacles & Procedural  & 0.307 & 0.948 & 2.278 \\
       & None  & 0.125 & 0.093 & 2.512 \\
       & Obs Avoid & \textbf{0.063} & \textbf{0.089} & \textbf{2.521} \\
\midrule
Flat  & Procedural & 0.127 & 0.371 & 2.320 \\
(Crowd)     & None & 0.087 & 0.082 & 2.374 \\
     & Agt Avoid & \textbf{0.013} & \textbf{0.071} & \textbf{2.402} \\
\bottomrule
\end{tabular}}
\end{center}
\vspace{-6mm}
\caption{Closed-loop animation results. Our system successfully follows waypoints and avoids collisions in a variety of terrains, and additional guidance improves performance.
}
\label{tab:anim-results}
\vspace{-5mm}
\end{table}

\cref{tab:anim-results} evaluates the animations from our system using \name with and without guidance in various settings: \textit{Random} is an assortment of smooth and rough slopes and stairs with varying difficulties, \textit{Obstacles} is a flat terrain with large obstacles, and \textit{Flat} is a flat terrain with pedestrians spawned in a crowd of 30.
For each setting, 600 rollouts of $10s$ are simulated across 30 characters with random bodies from AMASS~\cite{Mahmood2019AMASSAO}.
To put the difficulty of environments and discriminator rewards in context, we also include metrics when using the (terrain and obstacle unaware) \textit{Procedural} trajectory generation method used to train \animname. 

Our combined system performs well in the physically-simulated environment with \name providing easy-to-follow trajectories resulting in high-quality animations from \animname, as evaluated by the discriminator. 
Diffusion guidance can further improve failure rates, especially for avoiding agent collisions in dense crowds.
\cref{fig:anim-qual} shows some qualitative applications of our animation system and we highly encourage viewing the supplementary \textbf{video results} to qualitatively evaluate the motion quality. %
\cref{tab:value-results} shows the effect of using the learned value function from training \animname as a guidance loss for \name.
In each setting, adding value guidance in addition to waypoint guidance makes trajectories easier to follow, reduces failures, and improves the discriminator reward.
As a result, waypoint guidance error also improves.

\begin{table}
\begin{center}
\scalebox{0.7}{
\begin{tabular}{l c c c c c c}
\toprule
& \multicolumn{2}{c}{\textbf{Guide}} & \textbf{Waypoint} & \textbf{Fail} & \textbf{Traj Follow} & \textbf{Discrim} \\
\textbf{Terrain} & Waypoint & Value & Error & Rate & Error & Reward \\
\midrule
Random & $\surd$ & & 0.541 & 0.107 & \textbf{0.111} & 2.113 \\
       & $\surd$ & $\surd$ & \textbf{0.481} & \textbf{0.100} & 0.112 & \textbf{2.162} \\
\midrule
Obstacles & $\surd$ & & 1.065 & 0.220 & 0.138 & 2.552 \\
       & $\surd$ & $\surd$ & \textbf{0.929} & \textbf{0.178} & \textbf{0.113} & \textbf{2.609} \\
\midrule
Flat     & $\surd$ & & 0.248 & 0.063 & \textbf{0.084} & 2.555 \\
(Crowd)            & $\surd$ & $\surd$ & \textbf{0.175} & \textbf{0.053} & \textbf{0.084} & \textbf{2.607} \\
\bottomrule
\end{tabular}}
\end{center}
\vspace{-6mm}
\caption{Using the value function learned in RL training as guidance improves quality of trajectory following and robustness to varying terrains, obstacles, and other agents.
}
\label{tab:value-results}
\vspace{-6mm}
\end{table}

\vspace{-1mm}
\section{Discussion}
\label{sec:discussion}
\vspace{-1mm}
We have introduced a controllable trajectory diffusion model, a robust physics-based humanoid controller, and an end-to-end animation system that combines the two.
This represents an exciting step in being able to control the high-level behavior of learned pedestrian models, and opens several directions for future work.
First is improving the efficiency of sampling from trajectory diffusion models to make them real-time: \name currently takes $1$-$3s$ to sample for a single character, depending on the guidance used (see the supplement for full analysis). %
Recent work in diffusion model distillation~\cite{meng2022distillation} offers a potential solution.
In addition to high-level motion controllability, exploring how diffusion models can be extended to low-level full-body character control is an interesting next step.

{\small \paragraph{Acknowledgments} 
Davis Rempe was supported by an NVIDIA Graduate Fellowship.
The authors thank Ziyuan Zhong and the AV Research Group for helpful discussions on trajectory diffusion.
}

{\small
\bibliographystyle{ieee_fullname}
\bibliography{egbib}
}

\clearpage
\appendix
\section*{Appendices}
\cref{sec:supp-trace} and \cref{sec:supp-pacer} go over details of the \name and \animname models, respectively. 
\cref{sec:supp-expt-details} provides additional details of the experiments presented in the main paper, while \cref{sec:supp-results} gives additional results to supplement those in the main paper.
\cref{sec:supp-discussion} discusses limitations and future work in more detail.

\paragraph{Video Results}
Extensive video results are included on the \href{https://nv-tlabs.github.io/trace-pace}{project page}. 
We highly encourage readers to view them to better understand our method's capabilities.

\section{TRACE Details}
\label{sec:supp-trace}

In this section, we provide details on the \textbf{TRA}jectory Diffusion Model for \textbf{C}ontrollable P\textbf{E}destrians (\name) presented in Sec 3.1 of the main paper.

\subsection{Model Details}

\subsubsection{Denoising-Diffusion Formulation}
\paragraph{Input Representations}
In practice, the history trajectories of the ego pedestrian $\bx^\text{ego} = [ \bes_{t-T_p} \quad \dots \quad \bes_{t}]$ and $N$ neighboring pedestrians $X^\text{neigh} = \{\bx^\text{i} \}_{i=1}^{N}$ given as input to the diffusion model include more than just positions, heading, and speed.
In particular, each past state is 
$$\bes = [ x \quad y \quad h_x \quad h_y \quad v \quad l \quad w \quad p] \in \reals^8$$
where $(x,y)$ is the 2D position, $(h_x, h_y)$ is the 2D heading vector computed from the heading angle $\theta$, $v$ is the speed, $(l, w)$ is the 2D bounding box dimensions of the person, and $p \in \{0, 1\}$ indicates whether the person is present (visible) at that timestep or not (\eg, due to occlusions in real-world data). 
If a person is not visible at some step, \ie, $p = 0$, then the full state vector is zeroed out before being given to the diffusion model.
All trajectories are transformed into the local frame of the ego pedestrian at the current time step.

The rasterized map input $\map \in \reals^{H \times W \times C}$ is in bird's-eye view, and is cropped around the ego pedestrian and transformed into their local frame.
For all experiments $H = W = 224$ px at a resolution of 12 px/m. 
The map is cropped such that $14$ m to the front, left, and right of the ego are visible, and $\sim$$4.6$ m behind.
Each channel of $\map$ is a binary map with 1 indicating the presence of some semantic property. 
For example, in the \textit{ORCA} dataset, there are only two layers -- one for walkable area and one for obstacles. In nuScenes, there are seven layers representing lane, road segment, drivable area, road divider, lane divider, crosswalk, and sidewalk. Notice that the map for \name does not contain fine-grained height information as in the map for \animname. As such, \name is in charge of high-level obstacle avoidance while \animname factors in both obstacles and terrain. 

\paragraph{Denoising with Dynamics}
As discussed in the main paper, during denoising the future state trajectory is always a result of actions, \ie diffusion/denoising are on $\btau_a$, similar to \cite{zhong2022ctg}.
In detail, given an input noisy action sequence $\btau^k_a$ the denoising process is as follows: (\textbf{1}) compute the input state sequence $\btau^k_s = f(\bes_t, \btau^k_a)$ using the dynamics model $f$, (\textbf{2}) pass the full input trajectory $\btau^k = [\btau_s^k; \btau_a^k]$ to the denoising model to predict the clean action trajectory $\hat{\btau}^0_a$, (\textbf{3}) compute the output state trajectory $\hat{\btau}^0_s = f(\bes_t, \hat{\btau}^0_a)$, (\textbf{4}) if training, compute the loss in Eqn 3 of the main paper on the full output clean trajectory $\hat{\btau}^0 = [\hat{\btau}^0_s; \hat{\btau}^0_a]$.

We use a unicycle dynamics model for $f$~\cite{asl2022kinematics}.
Though humans are in theory more agile than the unicycle model, we find it regularizes predictions to be generally smooth, which is how pedestrians usually move and is amenable to being followed by an animation model.
Since our model requires actions as input, we compute these from the state-only input data through a simple inverse dynamics procedure.

\paragraph{Parameterization}
At each denoising step $k$, \name must predict the mean of the distribution used to sample the slightly less noisy trajectory for step $k-1$:
\begin{equation}
    p_\phi(\btau^{k-1} \mid \btau^k, C) := \normal (\btau^{k-1} ; \bmu_\phi(\btau^k, k, C), \boldsymbol{\Sigma}_k).
\end{equation}
There are three common ways to parameterize this prediction (we recommend \cite{nichol2021improved} for a full background on these formulations): (1) directly output $\bmu$ from the network, (2) output the denoised clean trajectory $\btau^0$, or (3) output the noise $\beps$ used to corrupt the clean trajectory.
\name uses (2), but the formulations are equivalent.
In particular, we can compute $\bmu$ from $\btau^k$ and $\btau^0$ using
\begin{equation}
\label{eqn:supp-mu-param}
    \bmu (\btau^0, \btau^k) := \frac{\sqrt{\bar{\alpha}_{k-1}} \beta_k}{1-\bar{\alpha}_k} \btau^0 + \frac{\sqrt{\alpha_k}\left(1-\bar{\alpha}_{k-1}\right)}{1-\bar{\alpha}_k} \btau^k
\end{equation}
where $\beta_k$ is the variance from the schedule (we follow \cite{janner2022diffuser,nichol2021improved} and use a cosine schedule), $\alpha_k := 1 - \beta_k$, and $\bar{\alpha}_k := \prod_{j=0}^k \alpha_j$.
Therefore, we can plug the output from \name $\hat{\btau}^0$ along with the noisy input $\btau^k$ into \cref{eqn:supp-mu-param} to get the desired next step mean $\bmu_\phi$.
We can also use the fact that $\btau^0$ is corrupted by
\begin{equation}
    \btau^k = \sqrt{\bar{\alpha}_k} \btau^0 + \sqrt{1-\bar{\alpha}_k} \beps
\end{equation}
where $\beps \sim \normal(\mathbf{0}, \mathbf{I})$ to compute $\beps$ from the output of \name:
\begin{equation}
    \beps = \frac{ \boldsymbol{\tau}^k - \sqrt{\bar{\alpha}_k} \boldsymbol{\tau}^0}{\sqrt{1-\bar{\alpha}_k}}.
\end{equation}
This allows the use of the classifier-free sampling strategy defined in Eqn 4 of the main paper, which requires mixing $\beps$ outputs from the conditional and unconditional models.

\begin{figure}[t]
\begin{center}
\includegraphics[width=0.85\linewidth]{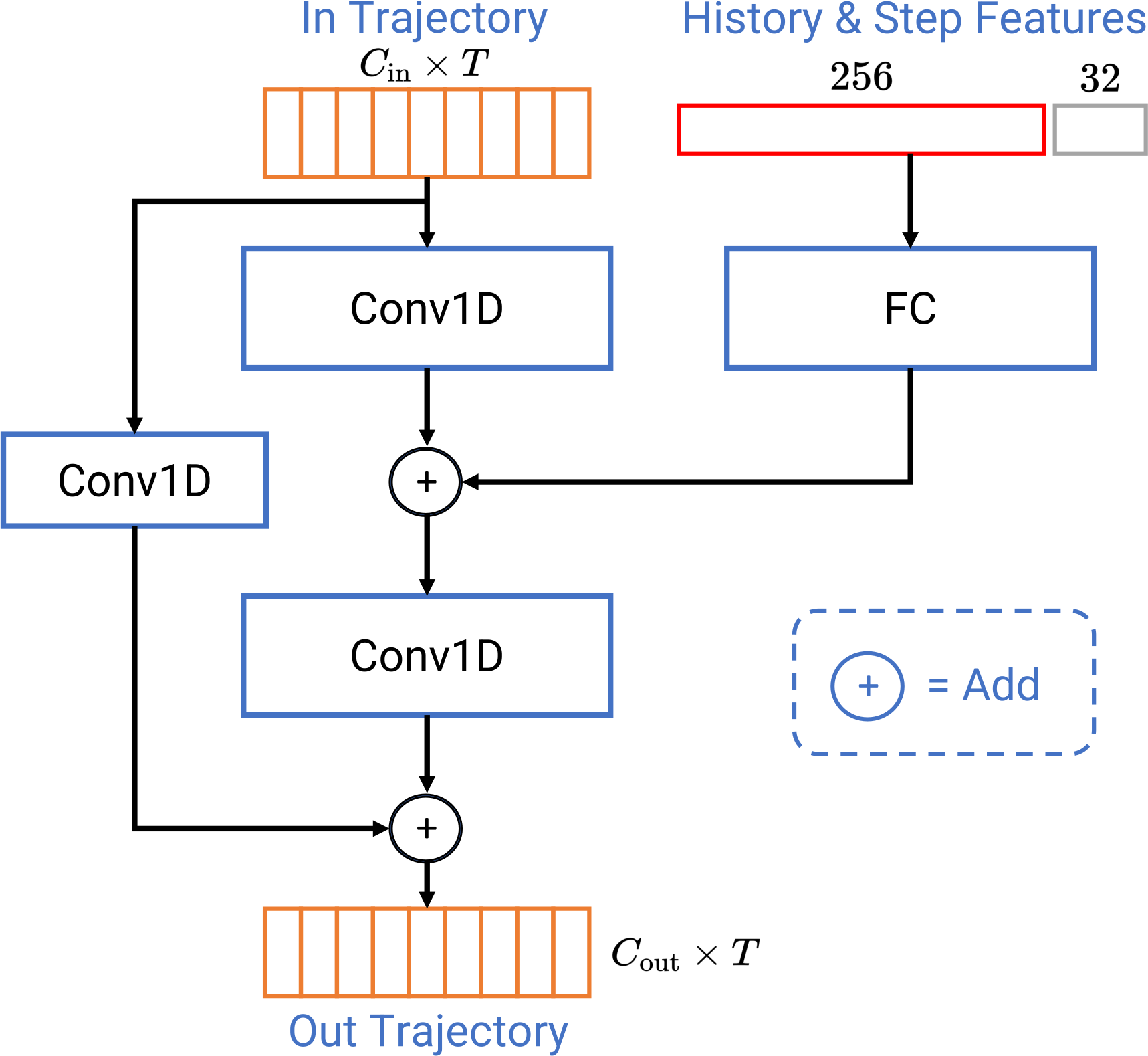}
\vspace{-3mm}
\end{center}
  \caption{Architecture of a single layer of denoising 1D U-Net.
  }
\vspace{-2mm}
\label{fig:supp-unet}
\end{figure}

\subsubsection{Architecture}
The denoising architecture is shown in Fig. 2 of the main paper.
At each denoising step $k$, the step index is processed with a positional embedding followed by a small MLP that gives a 32-dim feature. The map feature extractor uses the convolutional backbone of ResNet-18~\cite{he2016deep} as the encoder followed by a 2D U-Net~\cite{ronneberger2015u} decoder that leverages skip connections to layers in the encoder. For all experiments, the output feature grid $\bpsi$ is then $56 \times 56 \times 32$.

The ego $\bx^\text{ego}$ and neighbor $X^\text{neigh}$ history encoders operate on past trajectories $\in \reals^{T_p\times 8}$ that are flattened to be a single input vector.
The ego and all neighbor trajectories are processed by an MLP with 4 hidden layers giving a feature vector of size 128.
A different MLP is learned for ego and neighbor trajectories (\ie all neighbors are processed by the same MLP, which is different from the ego MLP).
Neighbor trajectory features are max-pooled to get a single \textit{interaction} feature.
The resulting ego and interaction features are finally jointly processed by another MLP with 4 hidden layers, giving a 256-dim feature summarizing the past trajectory context.

Note that the processing of input conditioning described thus far is only necessary to do \textit{once} before starting the denoising process. Only the denoising 1D U-Net needs to be run at every step.
At step $k$ of denoising, the 2D position at each timestep $t{+}i$ of the current noisy input trajectory $\btau^k$ is queried in the map feature grid to obtain a feature $\mathbf{g}_{t+i} = \bpsi(x_{t+i},y_{t+i}) \in \reals^{32}$. This query is done through bilinear interpolation of map features at the corresponding point. 
Over all timesteps, these form a feature trajectory $\mathbf{G} = [\mathbf{g}_{t+1} \dots \mathbf{g}_{t+T_f}]$ that is concatenated along the channel dimension with $\btau^k \in \reals^{T_f \times 6}$ (containing both actions and states) to get the full trajectory input to the denoising U-Net $[\btau^k; \mathbf{G}] \in \reals^{T_f \times 38}$.
Each layer of the U-Net also receives the concatenation of the past trajectory context and denoising step feature.

The architecture of a single U-Net layer is shown in \cref{fig:supp-unet}.
The input trajectory at each layer is first processed by a 1D convolution.
The input trajectory history and step index feature are projected to the same feature size, broadcast over the temporal dimension, and then added to the intermediate trajectory features.
Another convolution is performed before adding to the input trajectory in a residual fashion.
In the encoding part of the U-Net shown in Fig. 2 of the main paper, a 2$\times$ downsampling over the temporal dimension is performed between layers, while a 2$\times$ upsampling is done in the decoding part.
The encoder is three layers with output channels being 64, 128, and 256-dim.

\subsubsection{Training Details}
\name uses $K=100$ denoising steps in both training and testing.
Training uses a fixed learning rate of 2e-4 with the Adam optimizer~\cite{kingma2014adam} and runs for 40k iterations. 
TRACE trains on a 32 GB NVIDIA V100 GPU and takes $\sim$2 days on the ORCA dataset and $\sim$3 days on the mixed nuScenes+ETH/UCY data.

During training, the neighbor history and map conditioning are randomly dropped out with a 10\% probability. Note that these are dropped independently and that the ego history is never dropped.
In practice, to drop map conditioning, all pixels are filled with a $0.5$ value; this means that the model is aware that it ``does not know'' about the map context, it \textit{is not} simply fed an empty map with no obstacles (all zeros).
To drop neighbor conditioning, the neighbor history feature is zeroed out.
The same mechanism is used to train on ``mixed'' data with varying annotations, \eg, some data samples have no maps. In this case, a map is still given to the model but filled with $0.5$ value pixels.

\subsection{Guidance Details}

\subsubsection{Scene-Level Guidance}
Some guidance objectives are based on multi-agent interactions, \eg, agent collision avoidance and social groups.
In this case, we assume that all pedestrians in a scene can be denoised simultaneously in a batched fashion.
At each denoising step, the loss function is evaluated at the current trajectory prediction of all pedestrians and gradients are propagated back to each one for guidance.
This can be seen as sampling a scene-level future rather than a single agent future.
If we want to sample $M$ possible scene futures, we can draw $M$ samples from each agent and assume that the $m$th sample from each agent corresponds to the same scene sample. In other words, we compute the scene-level guidance by considering only the $m$th sample from each agent.

This multi-agent guidance slightly complicates the \textit{filtering} procedure described in Sec. 4 of the main paper whereby trajectory samples are strategically chosen to minimize the guidance loss.
In the case of a multi-agent objective, the trajectory that minimizes the guidance loss for one agent may not be globally optimal, so it is undesirable to filter the agents independently.
We instead do filtering at the scene level, similar to how guidance is computed: we compute the summed guidance loss across the $m$th sample from all agents and choose the \textit{scene-level sample} that minimizes this aggregate loss.

\begin{figure}[t]
\begin{center}
\includegraphics[width=0.95\linewidth]{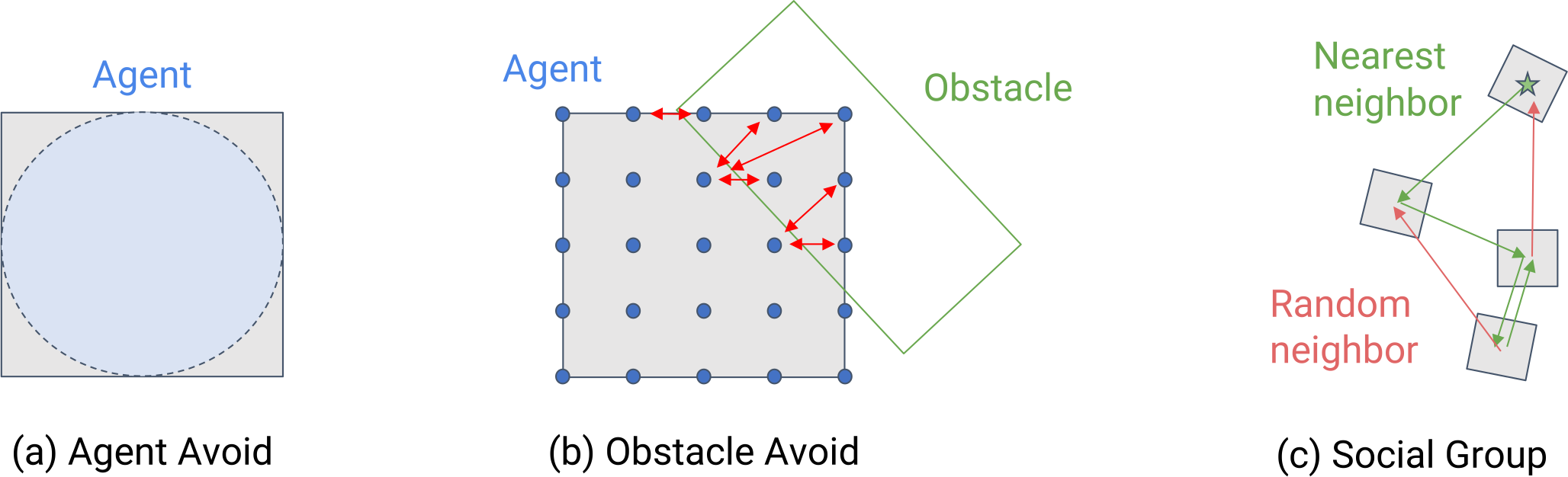}
\vspace{-3mm}
\end{center}
  \caption{Illustrations of various guidance objectives. See text for details.
  }
\vspace{-2mm}
\label{fig:supp-guidance}
\end{figure}

\subsubsection{Guidance Objectives}
\label{sec:supp-guides}

Next, we describe the different test-time guidance objectives (losses) that we have implemented for \name.
Objectives operate on the future state trajectory $\btau_s$ that starts at timestep $t+1$ and contains state $\bes_j$ at time $j$.

\paragraph{Agent Avoid and Social Distance}
We use the same agent collision penalty as in TrafficSim~\cite{suo2021trafficsim} and STRIVE~\cite{rempe2022strive} which approximates each agent with disks to efficiently and differentiably compute a collision loss.
For pedestrians, it is sufficient to use a single disk for each agent (see \cref{fig:supp-guidance}(a)).
With this approximation, collision detection is fast and the collision loss is computed based on the extent of disk overlap between agents.
It is easy to artificially inflate the size of the disk in order to implement a desired social distance between pedestrians.
Note that this is a multi-agent objective, so guidance is enforced at the scene level, as previously discussed.

\paragraph{Obstacle Avoid}
We extend the differentiable environment collision penalty introduced in STRIVE~\cite{rempe2022strive} to more robustly handle collision avoidance and provide more useful gradients.
The core idea is illustrated in \cref{fig:supp-guidance}(b); for each timestep where the pedestrian's bounding box is overlapping with an obstacle, we query a grid of points on the agent (in our experiments, this is $10{\times}10$) and define a loss with respect to points that are embedded in the obstacle.
For each embedded point, we compute the minimum distance $d_\text{min}$ to a non-embedded point on the agent and define the loss at that point as $\mathcal{L} = 1 - (d_\text{min} / b)$ where $b$ is the bounding box diagonal of the agent.
Summing the loss at all embedded points gives the total loss.

A subtle, but very important, implementation detail here is that the embedded points must be detached (\ie stop grad) before computing the loss. Intuitively, embedded points are treated as points on the obstacle, \textit{not} the agent. So when the loss is computed with respect to these points, it gives a meaningful gradient back to the non-embedded agent points which propagates back to the agent position and heading.

\paragraph{Local Waypoint at Specific Time}
This loss encourages an agent to be at a specific 2D goal waypoint $\bp_g = (x,y)$ at a specific time step $j$ that falls \textit{within} the planning horizon $T_f$ of the model.
It simply encourages the trajectory to be at that location at the timestep with the loss $\mathcal{L} = || \bes_j - \bp_g ||_2$.

\paragraph{Local Waypoint at Any Time}
This loss encourages an agent to be at a specific 2D goal waypoint $\bp_g = (x,y)$ at \textit{any} timestep that is \textit{within} the planning horizon $T_f$ of the model.
The loss is defined as
\begin{align}
    \mathcal{L} &= \sum_{j=t+1}^{t + T_f} \delta_j \cdot || \bes_j - \bp_g ||_2^2 \\
    \delta_j &= \frac{\exp(-|| \bes_j - \bp_g  ||)}{\sum_j \exp(-|| \bes_j - \bp_g ||)}.
\end{align}
Intuitively, it tries to minimize the distance from all points in the trajectory to the target location, but each timestep is weighted by $\delta_j$ which is the \textit{softmin} over the distances of each step from the waypoint.
The $\delta_j$ form a distribution over trajectory timesteps where states close to the waypoint will have higher probability and therefore be weighted more in the loss.

\paragraph{Global Waypoint at Specific Time}
This loss encourages an agent to be at a specific 2D goal waypoint $\bp_g = (x,y)$ at a specific timestep $j$ that falls \textit{beyond} the planning horizon $T_f$ of the model.
This is useful during closed-loop operation in which the agent should eventually reach the point, but at the current step $t$ is not within the planning horizon.
Intuitively, the loss encourages making enough progress toward the waypoint such that when it becomes in range, we can revert to the Local Waypoint loss and hit the target exactly at the desired time.

To do this, we would like to ensure that the future trajectory ends in a location such that the pedestrian can travel in a straight line at a ``preferred'' speed $v_\text{pref}$ and get to the waypoint on time.
Formally, the trajectory should be within a target distance defined as:
\begin{equation}
    d_\text{goal} = (j - t) \cdot dt \cdot v_\text{pref}
\end{equation}
where $dt$ is the step size of \name output (0.1 sec in our experiments).
Since the pedestrian may not be able to actually travel a straight line path (\eg in environments with obstacles), we incorporate an \textit{urgency} parameter $u \in [0, 1]$ that encourages getting there earlier and modifies the goal distance as
\begin{equation}
    \tilde{d}_\text{goal} = d_\text{goal} \cdot (1 - u).
\end{equation}
Then the loss with respect to this target distance is defined as
\begin{equation}
    \mathcal{L} = \text{ReLU}(|| \bes_{T_f} - \bp_g ||_2 - \tilde{d}_\text{goal})
\end{equation}
which penalizes the trajectory if the final state is not within the goal distance.

\paragraph{Global Waypoint at Any Time}
This loss encourages an agent to be at a specific 2D goal waypoint $\bp_g = (x,y)$ at \textit{any} timestep \textit{beyond} the planning horizon $T_f$ of the model.
To determine whether the goal waypoint is outside the current horizon, we check if the agent could progress along a straight line to the goal at a preferred speed $v_\text{pref}$ and reach the waypoint within the planning horizon.
If so, the loss reverts to the Local Waypoint loss.

If the waypoint is indeed beyond the planning horizon, the loss attempts to progress according to some urgency $u \in [0, 1]$.
To do this, we first compute the maximum distance that could be covered in the current horizon:
\begin{equation}
    d_\text{max} = T_f \cdot dt \cdot v_\text{pref}
\end{equation}
and use the urgency to get the goal distance we wish to cover over the horizon
\begin{equation}
    d_\text{goal} = u \cdot d_\text{max}.
\end{equation}
The loss is computed based on how much progress is made over the horizon:
\begin{align}
    \mathcal{L} &= \text{ReLU}( d_\text{goal} - d_\text{progress} ) \\
    d_\text{progress} &= || \bes_{t} - \bp_g ||_2 - || \bes_{t+ T_f} - \bp_g ||_2.
\end{align}

\paragraph{Social Groups}
This loss encourages groups of agents to travel together.
A social group is based on one \textit{leader} pedestrian that is not affected by the social group loss (via detach/stop grad); others in the group will tend to move with the leader.
Intuitively, we want each agent in the social group to maintain a specified social distance $d_\text{soc}$ to the closest agent also in the same social group.
Let $\psi$ be a map from one agent index to another, \eg $i = \psi(k)$ means agent $k$ in the social group is mapped to agent $i$.
Then the social group loss for agent $i$ at timestep $j$ in the future trajectory is
\begin{equation}
    \mathcal{L} = \left ( || \bes_j^i - \bes_j^{\psi(i)} ||_2 - d_\text{soc} \right)^2.
\end{equation}
As shown in \cref{fig:supp-guidance}, most of the time $\psi$ maps each agent to the closest agent in the group, but with some probability based on a cohesion parameter $c \in [0, 1]$ the mapping will be to a random agent in the group. 
So with a larger cohesion, agents in the group are all encouraged to be equidistant from each other, while with low cohesion agents will not closely follow the leader and connected components in the social group graph may break off and ignore others.

\paragraph{Learned Value Function}
Given a learned value function $V(\btau_s)$ that predicts the future rewards over a given trajectory, in general this guidance loss is simply $\mathcal{L} = \exp (-V(\btau_s))$. 
When \name is used with \animname, the value function is $V(v_t | \btau_s, \bs h_t, \bs o_t, \bs \beta)$; it takes in the current humanoid state $\bs h_t$, environmental feature $\bs o_t$, and humanoid shape $\bs \beta$, which are fixed throughout denoising.
\section{PACER Details}
\label{sec:supp-pacer}

In this section, we give details on the \textbf{P}edestrian \textbf{A}nimation \textbf{C}ontroll\textbf{ER} (\animname) presented in Sec 3.2.

\subsection{Implementation Details}

\paragraph{Humanoid State}
The state $\boldsymbol h_t$ holds joint positions $\boldsymbol{j^t} \in \mathbb{R}^{24 \times 3}$, rotations $\boldsymbol{q^t} \in \mathbb{R}^{24 \times 6}$, linear velocities $\boldsymbol{v^t} \in \mathbb{R}^{24 \times 3}$, and angular velocities $\boldsymbol{\omega^t} \in \mathbb{R}^{24 \times 3}$ all normalized w.r.t. the agent's heading and root position.
The rotation is represented in the 6-degree-of-freedom rotation representation. 
SMPL has 24 body joints with the root (pelvis) as the first joint, which is not actuated, resulting in an action dimension of $\bs{a}_t \in \mathbb{R}^{23 \times 3}$. No special root forces/torques are used.

\paragraph{Network Architecture} 
As mentioned in the main paper, the environmental feature $\bs o_t$ is a rasterized local height and velocity map of size $\bs o_t \in R^{64 \times 64 \times 3}$. The first channel is the terrain height map relative to the current humanoid root height, and the second \& third channels are the 2D linear velocities ($x$ and $y$ directions) in the egocentric coordinate system. The map corresponds to a 4m $\times$ 4m square area centered at the humanoid root, sampled on an evenly spaced grid. The trajectory $\btau_s \in R^{10 \times 2}$  consists of the 2D waypoints for the next 5 seconds sampled at 0.5 s intervals.

\begin{figure}[t]
\begin{center}
\includegraphics[width=0.95\linewidth]{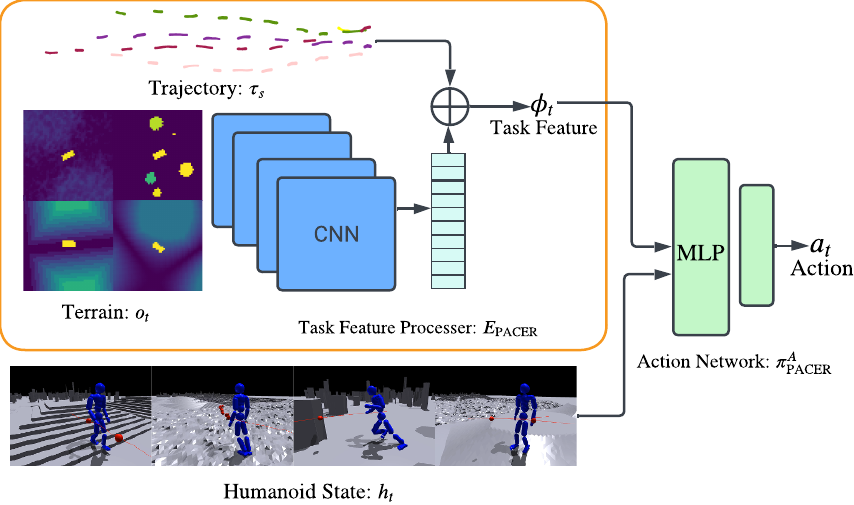}
\vspace{-6mm}
\end{center}
  \caption{The PACER policy network $\pi_{\text{PACER}}$ consists of a task feature processer $E_{\text{PACER}}(\bs \phi_t | \bs o_t, \bs \tau_s)$ and an action policy network $\pi^{\text{A}}_{\text{PACER}}(\phi_t, \bs h_t)$.
  }
\vspace{-4.5mm}
\label{fig:supp-pacer-arch}
\end{figure}

The architecture of $\pi_{\text{PACER}}$ can be found in \cref{fig:supp-pacer-arch}. Due to the high dimensionality of the environmental features $\bs o_t$, we separate the policy network into a task feature processer $E_{\text{PACER}}( \bs \phi_t | \bs o_t, \bs \tau_s)$ and an action network $\pi^{A}_{\text{PACER}}(\bs a_t | \bs \phi_t, \bs h_t, \bs \beta)$. The task feature processor transforms task-related features, such as environmental features $\bs o_t$ and trajectory $\bs \tau_s$ into a latent vector $\bs \phi_t \in \mathbb{R}^{256}$. Then, $\pi^{A}_{\text{PACER}}$ computes the action $\bs a_t$ based on the humanoid state $\bs h_t$, body shape $\bs \beta$, and $\bs \phi_t$. The overall policy network is then $ \pi_{\text{PACER}}(\bs a_t | \bs o_t, \bs h_t, \bs \beta, \bs \tau_s) \triangleq \pi^{A}_{\text{PACER}}(E_{\text{PACER}}(\bs o_t, \bs \tau_s), \bs h_t, \bs \beta) $.
$E_{\text{PACER}}$ is a four-level convolutional neural network with a stride of 2, 16 filters, and a kernel size of 4. $\pi^{A}_{\text{PACER}}$ is a standard MLP with ReLU activations. It has two layers, each with 2048 and 1024 units. The policy maps to the Gaussian distribution over actions $\pi_{\text{PACER}}(\bs a_t | \bs o_t, \bs h_t, \bs \beta, \bs \tau_s)  = \mathcal{N} (\mu (\bs o_t, \bs h_t, \bs \beta, \bs \tau_s), \Sigma)$ with a fixed covariance matrix $\Sigma$. Each action vector $\bs a_t \in \mathbb{R}^{23 \times 3}$ corresponds to the PD targets for the 23 actuated joints on the SMPL human body. The discriminator $\bs{D}(\bs{h}_{t-10:t}, \bs{a}_t)$ shares the same architecture as $\pi^{A}_{\text{PACER}}$, while the value function $V(v_t|  \bs o_t, \bs h_t, \bs \beta, \bs \tau_s)$ shares the same architecture as the policy $\pi_{\text{PACER}}$.

\subsection{Reward and Loss} 

\paragraph{Reward} 
Following AMP \cite{peng2021amp}, our policy $\pi_{\text{PACER}}$ is learned through goal-conditioned RL where the reward contains a task reward $r_t^{\bs \tau}$, a style reward $r_t^{\text{amp}}$, and an energy penalty $r_t^{\text{energy}}$. The style reward is computed by the discriminator $\bs{D}(\bs{h}_{t-10:t}, \bs{a}_t)$ based on 10 steps of aggregated humanoid state. We use the same set of observations, loss formulation, and gradient penalty as in AMP \cite{peng2021amp} to train our discriminator. Task reward $r_t^{\bs \tau}$ is a trajectory-following reward that measures how far away the humanoid's center $\bs c_t$ on the $xy$ plane is from the 2D trajectory: $ \exp (- 2 \times \| \bs c_t - \bs \tau_t\|^2)$. The energy penalty is expressed as $ -0.0005 \cdot \sum_{j \in \text { joints }}\left|\bs{\mu_j} \bs{\dot{q}_j}\right|^2 $ where $\bs{\mu_j}$ and $\bs{\dot{q}_j}$ correspond to the joint torque and the joint angular velocity, respectively.

\paragraph{Motion Symmetry Loss}
During our experiments, we noticed that asymmetric gaits emerge as training progresses. It manifests itself as ``limping" where the humanoid produces asymmetric motion, especially at a lower speed. This could be due to the small temporal window used in AMP (10 frames), which is not sufficient to generate symmetrical motion. Compared to AMP, we use a humanoid with more than double the degrees of freedom (69 vs 28), and the complexity of the control problem grows exponentially. This could also contribute to limping behavior as it becomes harder for the discriminator to discern asymmetric gaits. Thus, we utilize the motion symmetry loss proposed in \cite{Yu2018LearningSA} to ensure symmetric gaits. Specifically, we first design two functions $\Phi_s$ and $\Phi_a$ that can mirror the humanoid state and action along the character's sagittal plane. Symmetry is then enforced by ensuring that the mirrored states lead to mirrored actions: 

\vspace{-1mm}
\begin{equation}
\small
\begin{aligned}
    L_{\text{sym}}(\theta)&=\|\pi_\text{PACER}(\bs h_t, \bs o_t, \bs \beta, \btau_s) - \\
    &\Phi_a(\pi_\text{PACER}(\Phi_s(\bs h_t, \bs o_t, \bs \beta, \btau_s)))\|^2,  
\end{aligned}
\end{equation}

Notice that the motion symmetry loss is not a reward and is directly defined on the policy output. As the loss can be computed in an end-to-end differentiable fashion, we directly optimize this loss through SGD.

\subsection{Training}
Our training procedures closely follow AMP \cite{peng2021amp}, with notable distinctions in the motion dataset, initialization,  termination condition, terrain, and humanoids used. 
Training takes $\sim$3 days to converge on one NVIDIA RTX 3090.

\paragraph{Dataset} We use a small subset of motion sequences from the AMASS dataset \cite{Mahmood2019AMASSAO} to train our humanoid controller. Specifically, we hand-picked $\sim$200 locomotion sequences consisting of walking and turning at various speeds, as well as walking up and down stairs. These motions form the reference motion database and provide our AMP Discriminator $D(\bs h_t, \bs a_t)$ with ``real'' samples. 

\begin{figure}[t]
\begin{center}
\includegraphics[width=0.95\linewidth]{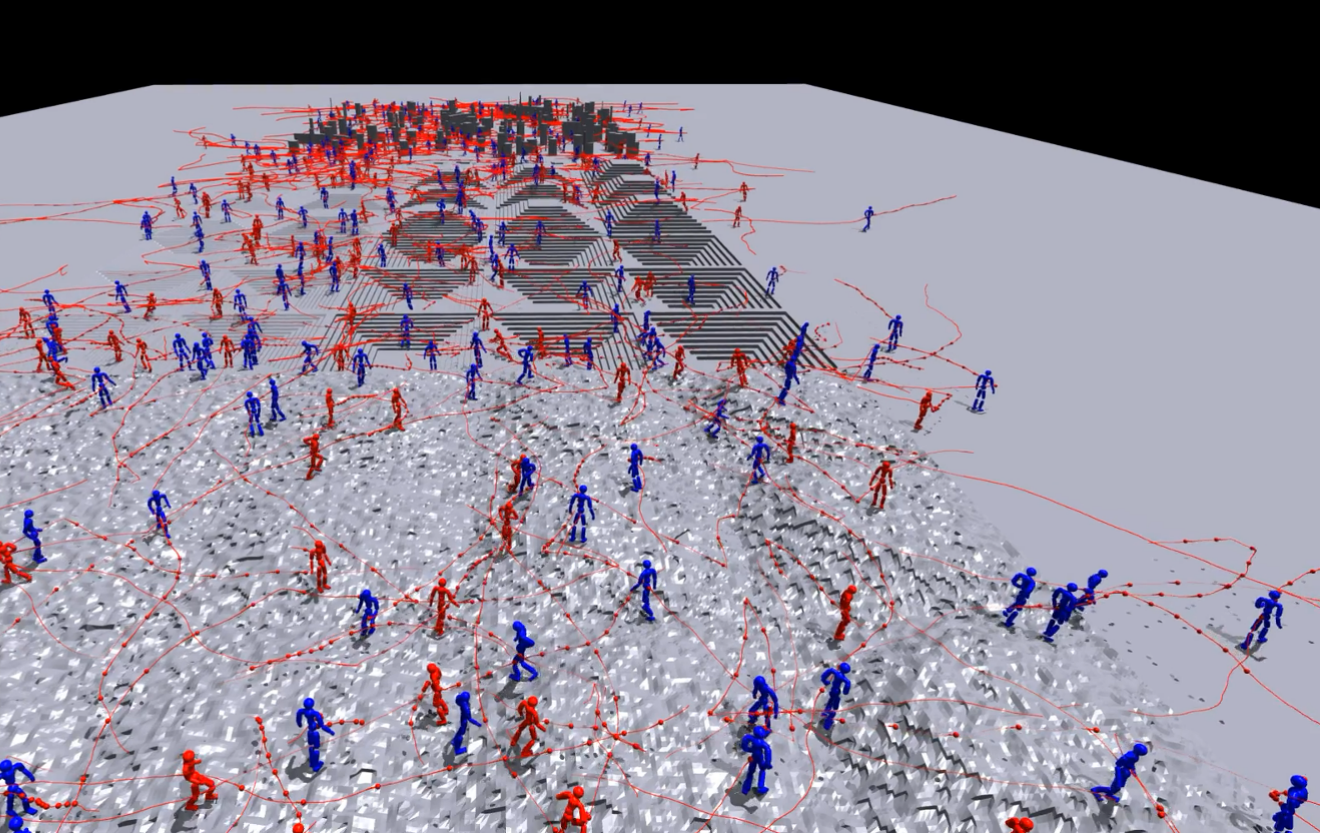}
\vspace{-6mm}
\end{center}
  \caption{During training, 2048 humanoids are simulated in parallel on our synthetic terrain. 
  }
\vspace{-4.5mm}
\label{fig:supp-sim}
\end{figure}

\paragraph{Initialization} To initialize our humanoids during training, we use reference state initialization \cite{2018-TOG-deepMimic} to randomly sample a body state $\bs h_0$. The initial root positions are randomly sampled from a ``walkable map'' that corresponds to all locations that can be used as a valid starting point (\eg not on top of obstacles). As we use NVIDIA's Isaac Gym, we create 2048 humanoids that are simulated simultaneously in parallel during training: see \cref{fig:supp-sim}.

\begin{figure}[t]
\begin{center}
\includegraphics[width=0.95\linewidth]{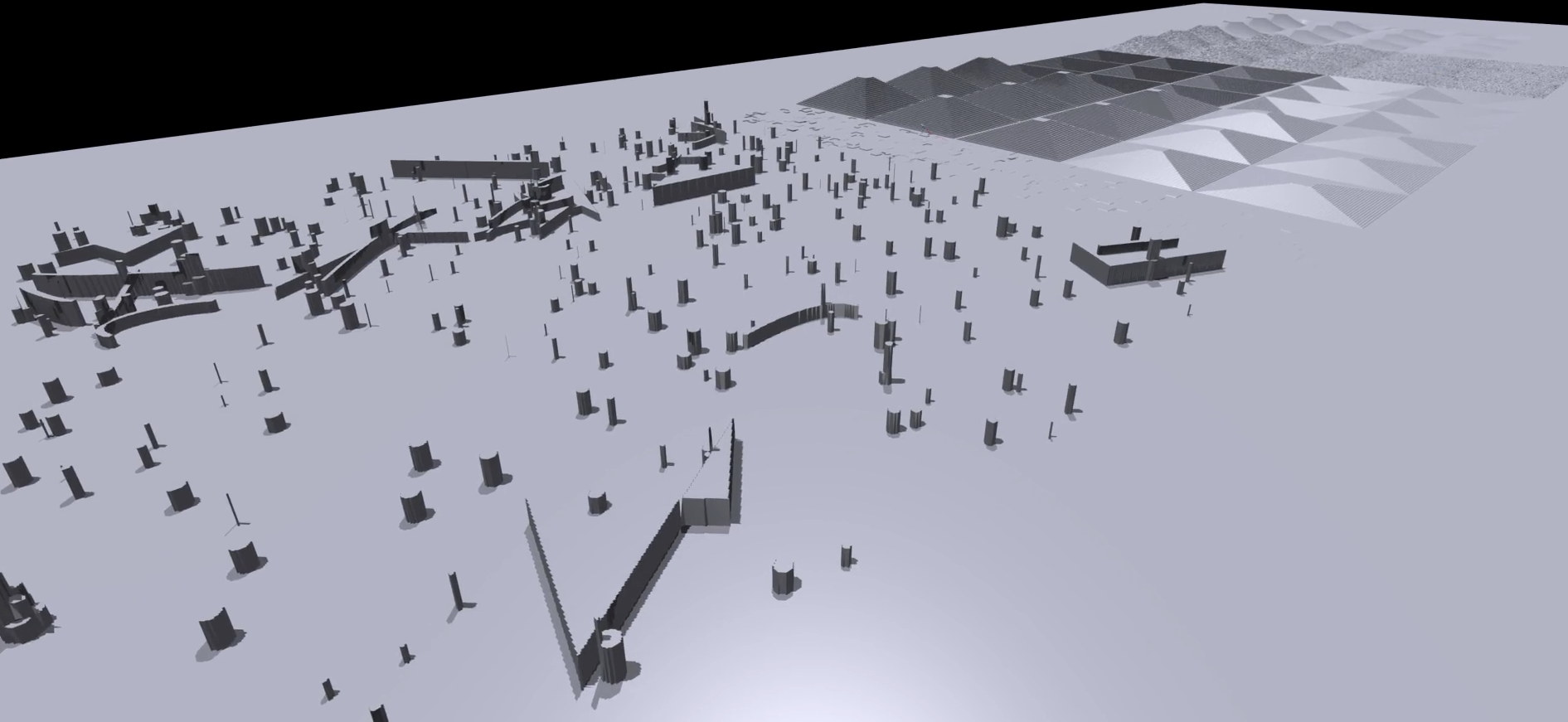}
\vspace{-6mm}
\end{center}
  \caption{Synthetic terrains used for training PACER. From left to right: obstacles, discrete terrains,  stairs (up), stairs (down), uneven terrains, and slopes. 
  }
\vspace{-4.5mm}
\label{fig:supp-terrains}
\end{figure}

\paragraph{Random terrain, trajectory, and body shape sampling}
To learn a model that can traverse diverse types of terrain that pedestrians may encounter in real life, we train our trajectory-following controller on a variety of different environments. Specifically, we follow ANYmal \cite{rudin2021learning} to create terrain curricula with varying difficulties to train our agents. Six types of terrain are created: slopes, uneven terrain, stairs (down), stairs (up), discrete, and obstacles. The terrains follow a gradual increase in difficulty, where we vary the slope angle, terrain unevenness, slope angle of stairs, and obstacle density, as shown in \cref{fig:supp-terrains}. 

Trajectory samples for training are generated procedurally: $\tau_s$ is randomly sampled by generating velocities and turn angles. We limit the velocity to be between [0, 3] m/s and the acceleration to be between [0, 2] m/$s^2$. 

To train with different body shapes, we extract all unique human body shapes from the AMASS dataset, which amount to 476 shapes (273 male and 200 female). We randomly sample (with replacement) 2048 body shapes to create humanoids at the beginning of the training process. To create reference humanoid states $\bs{\hat{h_t}}$ for the discriminator, we perform forward kinematics based on the sampled pose and the humanoids' kinematic tree. At the beginning of every 250 episodes, we randomly sample a new batch of pose sequences from the motion dataset and create new reference humanoid states. In this way, we obtain reference states of diverse body types and motions.

\paragraph{Termination condition}
To speed up training, we employ early termination \cite{2018-TOG-deepMimic} and terminate the episode if there is a collision force greater than 50 N on the humanoid body, with either the scene or other humanoids. The ankles and foot joints are exceptions to this rule, as they are in contact with the ground. This condition also serves as a fall detection mechanism, as falling will involve a collision force from the ground. Notice that this termination condition encourages the humanoid to avoid obstacles and other humanoids since a collision will trigger an early termination.

\section{Experimental Details}
\label{sec:supp-expt-details}

In this section, we include details of the experiments presented in Sec 4 of the main paper.

\subsection{Dataset Details}
The \textit{ORCA} dataset contains two distinct subsets, \textit{ORCA-Maps} and \textit{ORCA-Interact}. ORCA-Maps is generated with up to 10 pedestrians and 20 obstacles in each scene. This contains many obstacle interactions, but fewer agent-agent interactions. ORCA-Interact has up to 20 pedestrians, but no obstacles, and therefore has no map annotations. Each data subset contains 1000 scenes that are $10s$ long, and we split them 0.8/0.1/0.1 into train/val/test splits.
The map annotations in the ORCA dataset contain two channels, one representing the walkable area and one representing obstacles.
The bounding box diameter for every agent is fixed to $0.8m$.

In nuScenes~\cite{caesar2020nuscenes}, there are seven map layers representing the lane, road segment, drivable area, road divider, lane divider, crosswalk, and sidewalk. 
The bounding box diameters are given by the dataset.
We follow the official trajectory forecasting benchmark for scenes in the train/val/test splits.
For ETH/UCY~\cite{pellegrini2009you,lerner2007crowds}, we use the official training splits of each contained dataset for training.

All trajectory data in all datasets is re-sampled to 10 Hz for training and evaluation of \name. 

\subsection{Guidance Metrics}
Here, we define the \textit{Guidance Error} for each of the objectives evaluated in Sec 4.1 and 4.2 of the main paper.

\paragraph{Obstacle Avoid}
This is the obstacle collision rate as defined below.

\paragraph{Agent Avoid}
This is the agent collision rate as defined below.

\paragraph{Waypoint and Perturbed Waypoint}
If the objective is to reach a waypoint at a specific timestep, this is simply the distance of the agent from the target waypoint at that specified timestep (in meters). Otherwise, if the objective is to reach at \textit{any} timestep, the error is the minimum distance between the agent and the goal waypoint across the entire trajectory.

\paragraph{Social Groups}
For each pedestrian in the group, we measure the mean absolute difference between the specified social distance $d_\text{soc}$ (see \cref{sec:supp-guides}) and the distance to the closest neighbor in the same social group.

\paragraph{Multi-Objective}
For the multi-objective guidance presented in Tab 1 of the main paper (Waypoint + Avoidance), the reported guidance error is the waypoint error since obstacle and agent collision rates are already reported in other columns.

\subsection{Other Metrics}
Next, we describe in more detail the metrics used to evaluate the standalone \name model in Sec 4.1 and 4.2 of the main paper.

\paragraph{Obstacle Collision Rate}
Measures the average fraction of time that an agent (as represented by a single disk) is overlapping with an obstacle on the map within a rollout.
Note that a disk is used to represent each agent because this is how they are represented in the ORCA simulator, so the ground truth data contains no collisions using this representation.

\paragraph{Agent Collision Rate}
Measures the average fraction of agents involved in an agent-agent collision within each scene rollout. This again uses the disk representation of each agent. 

\paragraph{Realism (EMD)}
Compares the histogram of statistics over the entire test set between generated and ground truth trajectories.
This is done for velocity, longitudinal acceleration, and lateral acceleration.
In particular, the statistics at each timestep of the test set are aggregated together into a histogram. The histogram is then normalized such that it sums to 1.
The earth mover's distance (EMD) between the ground truth and generated histograms is then computed\footnote{using \href{https://github.com/wmayner/pyemd}{pyemd}} and reported.
Note that this metric is computed wrt \textit{the dataset being evaluated on}. For example, in Sec 4.1 of the main paper, even though \name is trained on both ORCA-Maps and ORCA-Interact, the metric is only computed for ORCA-Maps since this is the test data.

\paragraph{Realism (Mean)}
Measures the average longitudinal and lateral acceleration within a generated trajectory in $m/s^2$.
Similar metrics are commonly used in the vehicle planning literature~\cite{wang2021advsim} as a proxy for how \textit{comfortable} a ride is. In our case of pedestrian motion, this is still relevant since people tend to move in smooth motions without sudden changes in speed or direction.

\subsection{VAE Baseline Details}
We adapt a conditional VAE model similar to the idea of STRIVE~\cite{rempe2022strive} for controlling trajectories through latent space optimization. We adapt the VAE design to our setting.

\paragraph{Architecture} 
The architecture operates in an agent-centric manner as in \name.
It is a fairly standard conditional VAE (CVAE) where the conditioning (map and past trajectories) is processed into a single conditioning vector $\bc$ that is given to the decoder. At training time, the decoder also takes in a latent vector $\bz$ from the encoder (posterior), while at test time the latent vector is sampled from the prior $p(\bz) = \normal(\bz; \mathbf{0}, \mathbf{I})$.
To make the methods comparable, the map conditioning is encoded with the same ResNet-18 backbone that \name uses; ego and neighbor past trajectories also use the same architecture as \name.
Since the model is agent-centric rather than scene-centric, the decoder $D$ is simply an MLP that maps the conditioning and sampled latent to an output action trajectory (instead of a graph network as in STRIVE) as $\btau_a = D(\bz, \bc)$.
In all experiments, the latent dimension is 64, while the conditioning feature vector is 256-dim.

\paragraph{Training}
Training is done using a standard VAE loss consisting of a reconstruction and a KL divergence term. 
The KL term is weighted by 1e-4.
The model is trained with the same batch size as \name (400) and for the same number of iterations (40k) with a learning rate of 2e-4.

\paragraph{Test-Time Optimization}
The idea of test-time optimization is to search for a latent vector that is likely under the prior (\ie represents a plausible future trajectory) but also meets the desired guidance objective.
Concretely, the optimization objective is
\begin{equation}
\label{eqn:supp-vae-guide}
    \min_{z} \alpha \guide (D(\bz, \bc)) - \log p(\bz)
\end{equation}
where $\guide$ is a guidance loss as described in the main paper and $\alpha$ balances the prior term with the guidance loss.
Optimization is performed with Adam~\cite{kingma2014adam} using a learning rate of $0.02$.
For experiments in Sec 4.1 of the main paper, optimization uses 100 iterations (same as the number of diffusion steps $K$).
For Sec 4.2, the iteration budget is increased to 200 to accommodate more difficult out-of-distribution objectives.

\paragraph{Discussion on VAE Comparison} The VAE with test-time optimization is generally a very strong baseline.
Given a large enough compute budget, the optimization can usually faithfully meet the desired objective.
However, the number of optimization iterations needed to meet an objective can be large; \eg in Sec 4.2 it requires twice the number of diffusion steps, making it slower than \name.
Moreover, when optimizing for a long time to closely meet objectives, the diversity of optimized samples becomes low as they converge to similar minima (Fig. 4 in the main paper). This is due to the prior term in \cref{eqn:supp-vae-guide}, which always drives the trajectory towards the mean.

\subsection{Additional Experiment Details}
Finally, we include miscellaneous details of the setup for each experiment in Sec. 4 of the main paper.

\paragraph{Augmenting Crowd Simulation (Sec 4.1)} 
For the no guidance rows in Tab 1, we actually run the evaluation three times and report the averaged metrics. This is because when there is no guidance, no filtering is performed, so a random sample is chosen. Running with several random samples gives a more faithful evaluation of performance.
In this experiment, \name uses $w=0.0$ for classifier-free sampling. 20 samples are drawn and guided from the model for each pedestrian before filtering.
The weighting $\alpha$ for each guide (in Eq. 6 of the main paper) is tuned manually to meet objectives while maintaining realistic trajectories.
The \textit{Waypoint} guidance used in Tab 1 is the \textit{Local Waypoint at Any Time} introduced in \cref{sec:supp-guides}.
The agent avoidance guidance uses an additional social distance buffer of $0.2m$.

\paragraph{Real-world Data Evaluation (Sec 4.2)}
In this experiment, 10 samples are drawn from the model before filtering and the waypoint guidance is \textit{Global Waypoint at Any Time} since the model operates in a closed loop for longer than the planning horizon.
This waypoint guidance uses an urgency of $u=0.7$ and a preferred speed of $v_\text{pref} = 1.25 m/s$.
The perturbed waypoint objective randomly perturbs the target ground truth waypoint with Gaussian noise with a standard deviation of $2m$.
The social group guidance uses $d_\text{soc} = 1.5$ and cohesion $c=0.3$. In each nuScenes scene, social groups are determined heuristically by forming a scene graph where edges are present if two pedestrians are within $3m$ of each other and moving in a similar direction (velocities have a positive dot product): the connected components of this graph with more than one agent form the social groups.

\paragraph{Controllable Pedestrian Animation (Sec 4.3)}
In this experiment, 10 samples are drawn from the model before filtering.
Waypoint guidance uses \textit{Global Waypoint at Specific Time} with the waypoint randomly placed at a reasonable distance ($[7, 12]$ meters in front of the user and up to $5$ meters to either side) 9 sec in the future.
Agents are initialized in a standing pose with a uniform random initial root velocity in $[1.2, 2.0] m/s$.
\section{Supplementary Results}
\label{sec:supp-results}

In this section, we include additional experimental results omitted from the main paper due to page limits.

\subsection{Qualitative Results}
Extensive qualitative video results for both \name and \animname are provided \textbf{on the supplementary webpage}.

\begin{table*}[t]
\begin{center}
\scalebox{0.85}{
\begin{tabular}{lcc | c c | c c | c c c}
\toprule
 & \textbf{History} & \multicolumn{1}{c}{\textbf{Map}}  & \multicolumn{2}{c}{\textbf{Accuracy}} & \multicolumn{2}{|c}{\textbf{Collision Rate}} & \multicolumn{3}{|c}{\textbf{Realism} \small{(EMD)}} \\
 \textbf{Method} & Input & Feature & ADE & FDE & Obstacle & Agent & Vel & Lon Acc  & Lat Acc \\
\midrule
VAE~\cite{rempe2022strive} & States & Global & 0.340 & 0.774 & 0.062 & 0.115 & 0.041 & 0.038 & 0.039 \\
\name-Raster & Raster & Global & 0.337 & 0.808 & 0.052 & 0.100 & 0.027 & \textbf{0.013} & \textbf{0.014} \\
\name-Global & States & Global & \textbf{0.280} & \textbf{0.686} & 0.056 & \textbf{0.094} & \textbf{0.022} & \textbf{0.013} & 0.016 \\
\midrule
\name & States & Grid & 0.318 & 0.757 & \textbf{0.046} & 0.110 & 0.028 & 0.020 & 0.020  \\
\bottomrule
\end{tabular}}
\end{center}
\vspace{-6mm}
\caption{No guidance evaluation on \textit{ORCA-Maps} dataset. Ablation on architecture design choices.
}
\label{tab:supp-trace-arch-ablation}
\vspace{-1mm}
\end{table*}
\begin{table*}[t]
\begin{center}
\scalebox{0.85}{
\begin{tabular}{lc | c c | c c | c c c}
\toprule
 & \textbf{Drop}  & \multicolumn{2}{c}{\textbf{Accuracy}} & \multicolumn{2}{|c}{\textbf{Collision Rate}} & \multicolumn{3}{|c}{\textbf{Realism} \small{(EMD)}} \\
\textbf{Train Data} & Rate & ADE & FDE & Obstacle & Agent & Vel & Lon Acc  & Lat Acc \\
\midrule
ORCA-Maps & 10\% & 0.351 & 0.819 & \textbf{0.040} & 0.112 & 0.030 & 0.023 & 0.024 \\
Mixed & 0\% & \textbf{0.303} & 0.719 & 0.042 & 0.123 & 0.028 & \textbf{0.019} & \textbf{0.020}  \\
Mixed & 5\% & 0.307 & \textbf{0.712} & \textbf{0.040} & \textbf{0.108} & \textbf{0.024} & 0.020 & 0.023 \\
\midrule
Mixed & 10\% & 0.318 & 0.757 & 0.046 & 0.110 & 0.028 & 0.020 & \textbf{0.020} \\
\bottomrule
\end{tabular}}
\end{center}
\vspace{-6mm}
\caption{No guidance evaluation on \textit{ORCA-Maps} dataset. Ablation on training routine.
}
\label{tab:supp-trace-train-ablation}
\vspace{-3mm}
\end{table*}
\begin{figure}[t]
\begin{center}
\includegraphics[width=0.95\linewidth]{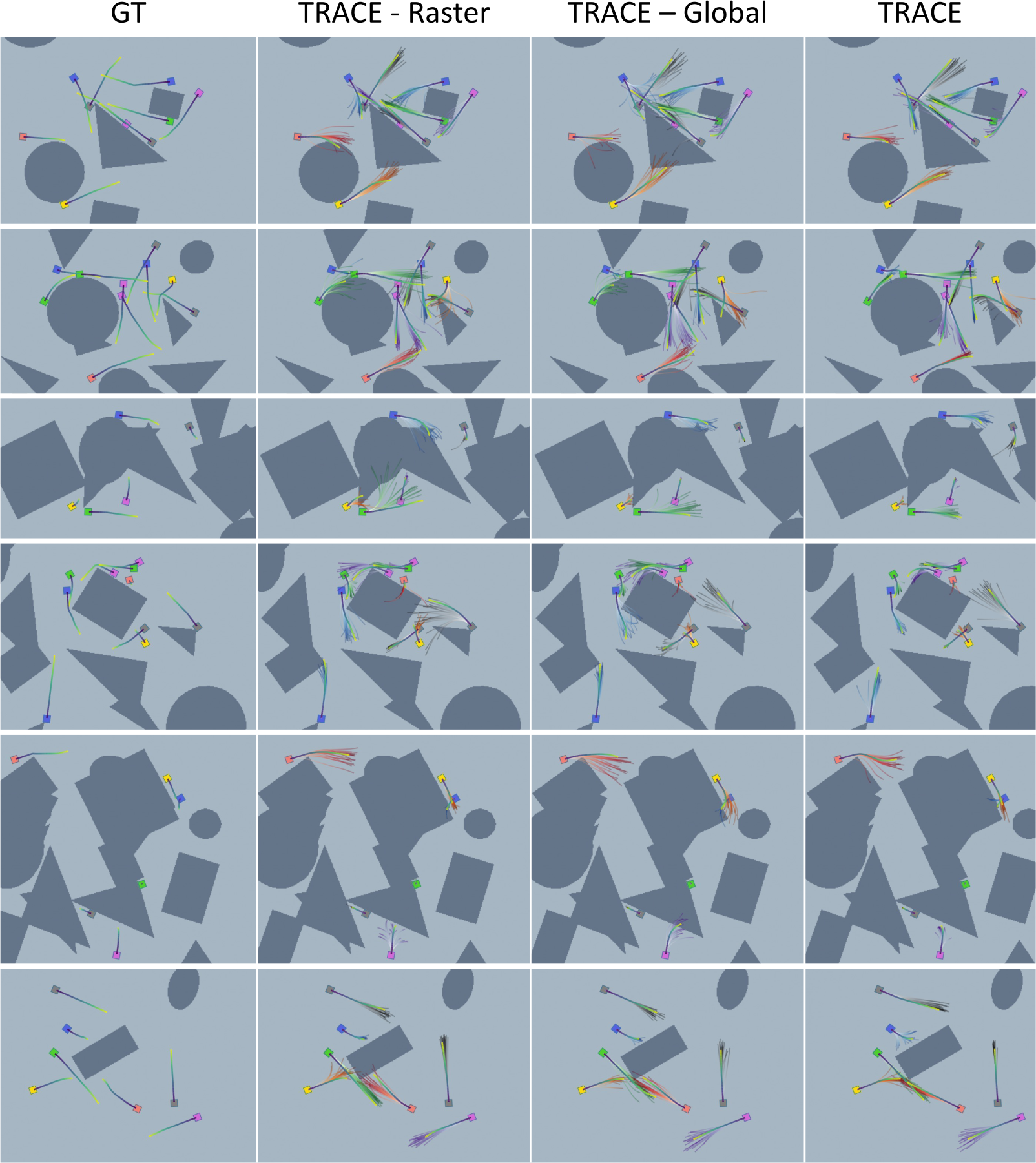}
\vspace{-6mm}
\end{center}
  \caption{Random sampling with no guidance from different \name architecture ablations. Using the learned feature map has apparent benefits in subtle interaction with obstacles. 20 random samples are visualized for each pedestrians with the (arbitrarily) chosen plan in bold.
  }
\vspace{-4.5mm}
\label{fig:supp-arch-compare}
\end{figure}

\subsection{\name -- No Guidance Ablation Study}
The focus of our work is on controllability using guidance. 
However, it is still desired that the model performs well even when guidance is not used.
In particular, random samples from the model should be robust (avoid collisions), realistic (similar to the training data distribution), and accurate (capture the ground truth future trajectory).
To this end, we evaluate our architecture and training approach compared to baselines and ablations while using no guidance.
We evaluate in the open-loop $5s$ rollout setting on ORCA-Maps as used in Sec. 4.1 of the main paper.
Two additional metrics common in trajectory forecasting are evaluated: the average displacement error (ADE) and the final displacement error (FDE)~\cite{yuan2021agentformer}. For a trajectory sample defined from timesteps 1 to $T$, these are defined as $\text{ADE}=\frac{1}{T} \sum_{t=1}^T || \hat{\mathbf{y}}_t-\mathbf{y}_t ||_2$ and $\mathrm{FDE}= || \hat{\mathbf{y}}_{T}-\mathbf{y}_{T} ||_2$ with $\hat{\mathbf{y}}$ the sample from the model and $\mathbf{y}$ the ground truth.

First, we evaluate the \name architecture, which uses a feature grid to condition denoising on the map input.
An alternative way to condition trajectory generation is to encode the map into a single (``global'') feature vector using a convolutional backbone. This global feature can then be given in the same way as the past trajectory features.
The \textit{VAE} baseline and \textit{\name-Global} ablation do this using a ResNet-18 backbone.
The \textit{\name-Raster} ablation is similar to CTG~\cite{zhong2022ctg}, which rasterizes both the map \textit{and} agent histories and encodes them into a single global feature instead of encoding the trajectory states separately. 
\cref{tab:supp-trace-arch-ablation} shows the results comparing these methods. In this experiment, we take 20 samples from each model and evaluate the one that is closest to the ground truth wrt the ADE.
We see that using the feature grid map provides the lowest obstacle collision rate while maintaining competitive accuracy and realism. 
As qualitatively shown in \cref{fig:supp-arch-compare}, the use of the feature grid gives local cues to the model to inform subtle obstacle interactions and avoid collisions.
Though agent collisions are slightly worse with the grid map, no model does particularly well, and exploring improved agent-agent interactions is an important direction for future work.

In the main paper, we discuss how mixed training data and classifier-free sampling (\ie training with random dropping on conditioning) are important to enable flexibility for guidance. To ensure this training approach does not negatively affect base model performance without guidance, we compare to (1) an ablation that uses the ORCA-Maps data only to train (rather than a mix of ORCA-Maps+ORCA-Interact) and (2) ablations that use varying levels of dropping.
\cref{tab:supp-trace-train-ablation} shows results, where again the sample closest to ground truth is evaluated.
Interestingly, training with mixed data allows for increased accuracy compared to training only on the ORCA-Maps dataset. 
Increasing the drop probability past 5\% has very little effect on performance and comes with the added benefit of using classifier-free sampling to get flexible guidance at test time.

\begin{table}[t]
\begin{center}
\scalebox{0.9}{
\begin{tabular}{c | c c | c c c}
\toprule
   & \multicolumn{2}{|c}{\textbf{Collision Rate}} & \multicolumn{3}{|c}{\textbf{Realism} \small{(EMD)}} \\
$w$ & Obstacle & Agent & Vel & Lon Acc  & Lat Acc \\
\midrule
0.0 & 0.051 & 0.131 & 0.019 & 0.012 & 0.014  \\
0.3 & 0.051 & 0.130 & 0.027 & 0.008 & 0.010 \\
0.5 & 0.050 & 0.132 & 0.029 & 0.008 & 0.009 \\
0.7 &  0.050 & 0.132 & 0.033 & 0.009 & 0.008  \\
1.0 & 0.049 & 0.130 & 0.040 & 0.010 & 0.009  \\
2.0 &  0.051 & 0.132 & 0.063 & 0.017 & 0.015  \\
3.0 & 0.052 & 0.138 & 0.087 & 0.025 & 0.022 \\
4.0 & 0.051 & 0.145 & 0.102 & 0.033 & 0.028  \\
\bottomrule
\end{tabular}}
\end{center}
\vspace{-6mm}
\caption{Classifier-free sampling analysis on \textit{ORCA-Maps} dataset with no guidance.
}
\label{tab:supp-class-free-no-guide}
\vspace{-4mm}
\end{table}
\begin{table}[t]
\begin{center}
\scalebox{0.85}{
\begin{tabular}{c | c | c c}
\toprule
   & \textbf{Waypoint} & \multicolumn{2}{|c}{\textbf{Realism} \small{(Mean)}} \\
$w$ & Error & Lon Acc  & Lat Acc \\
\midrule
0.0 & 1.129 & 0.233 & 0.218 \\
-0.3 & 0.972 & 0.213 & 0.199 \\
-0.5 &  0.802 & 0.212 & 0.204 \\
-0.7 & 0.670 & 0.240 & 0.233  \\
-1.0 &  0.546 & 0.345 & 0.348 \\
\bottomrule
\end{tabular}}
\end{center}
\vspace{-6mm}
\caption{Classifier-free sampling analysis on nuScenes dataset using perturbed waypoint guidance.
}
\label{tab:supp-class-free-guide}
\vspace{-3mm}
\end{table}
\begin{figure}[t]
\begin{center}
\includegraphics[width=0.95\linewidth]{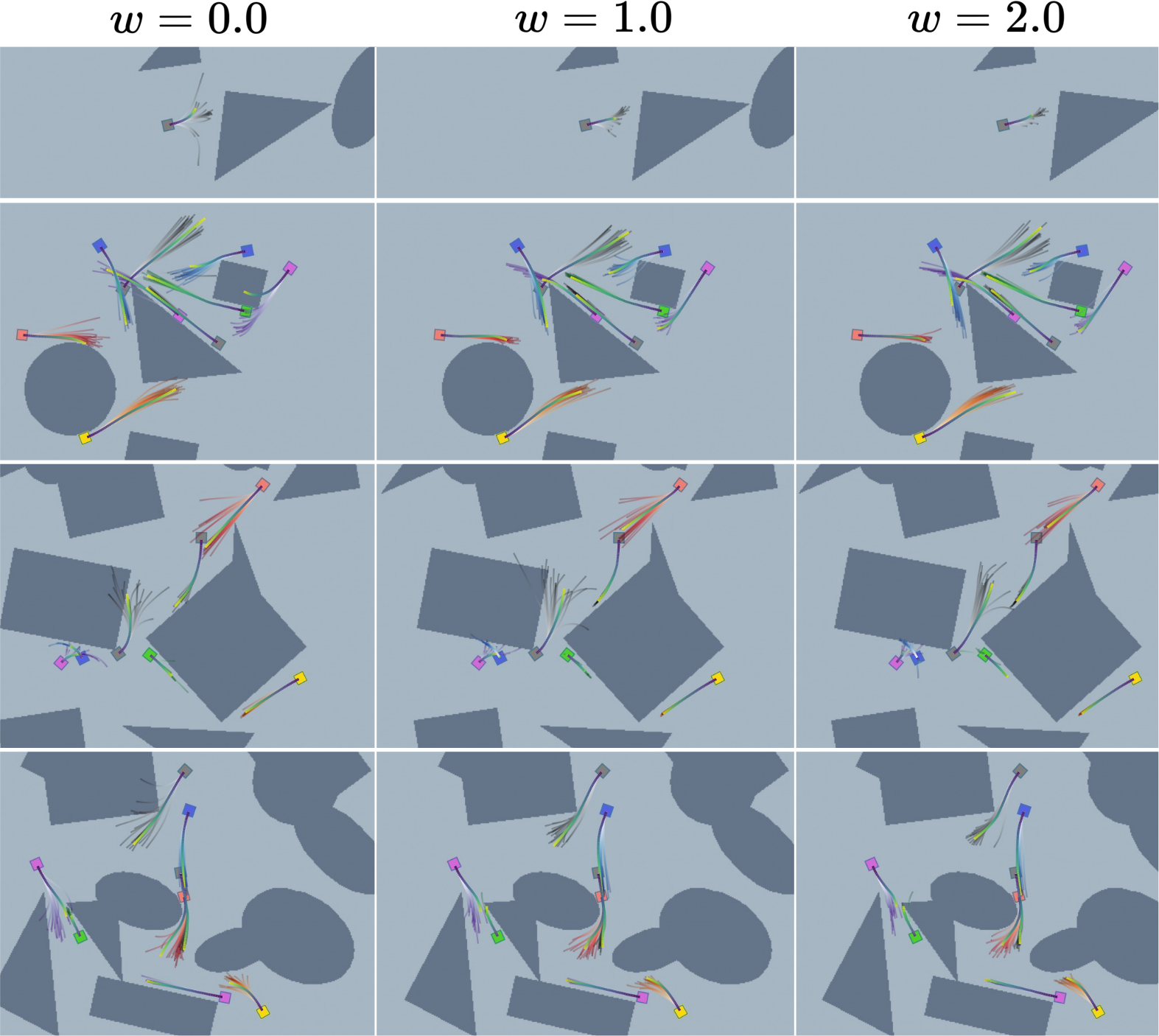}
\vspace{-6mm}
\end{center}
  \caption{Sampling using increasing classifier-free weights $w$. 20 samples are visualized for each pedestrian. Larger $w$ tends to emphasize collision avoidance and reduces the variance of the sampled trajectory distribution, especially for pedestrians near obstacles where conditioning has a large effect on motion.
  }
\vspace{-2mm}
\label{fig:supp-orca-w-sweep}
\end{figure}

\subsection{\name -- Effect of Classifier-Free Sampling}
Next, we examine how weight $w$ affects model performance when using classifier-free sampling both with and without guidance.
First, we analyze the effect when evaluating on the ORCA-Maps dataset with no guidance in the open-loop setting (like Sec 4.1 of the main paper).
In this case, we evaluate $w \geq 0$ which increases emphasis on the input conditioning to the model.
Quantitative results are shown in \cref{tab:supp-class-free-no-guide}: for each value of $w$, the evaluation is run 3 times with different random samples and the metrics are averaged.
For $w \in [0, 1]$, the collision rates and acceleration realism remain similar or slightly improve, which we expect since input conditioning such as the obstacle map is emphasized. For $w{>}1$, the guidance tends to be too strong and the trajectory samples are almost deterministic.
Though the quantitative difference is not large as $w$ increases, in \cref{fig:supp-orca-w-sweep} we see that increasing does have a considerable qualitative effect.

Second, we look at results on the nuScenes dataset using perturbed waypoint guidance in the closed-loop setting (the same as in Sec. 4.2 of the main paper). In Sec. 4.2 of the main paper, we saw that using $w<0$ improves susceptibility to guidance and allows the model to achieve out-of-distribution objectives. This is further confirmed in \cref{tab:supp-class-free-guide}. 
We see that the smaller the $w$, the better the waypoint reaching error. However, for $w<-0.5$ the mean accelerations of pedestrians start to deviate more from those observed in the ground truth nuScenes data, as the model is capable of producing more extreme trajectories to reach waypoints.

\subsection{\animname -- Ablation Study}
In this experiment, we demonstrate the importance of multiple design decisions in the \animname model.
First, we choose to make the animation controller agent-aware by including neighboring pedestrians in the heightmap given to the model. 
For comparison, we train a model that is agent \textit{unaware}, \ie, the input height map only contains obstacles.
As seen in the top half of \cref{tab:supp-pacer-ablation}, even though \name is already agent-aware, having \animname endowed with awareness is highly beneficial. Both with and without agent avoidance guidance on \name, the agent-aware model greatly improves the collision rate.

Second, we evaluate whether making \animname body-aware is necessary, \ie if it needs to take the parameters of the SMPL body $\beta$ shape as input, as it is already training in simulation with a variety of body shapes.
The bottom half of \cref{tab:supp-pacer-ablation} shows that while traversing random terrains, body awareness helps to improve the failure rate when no guidance is used. When waypoint guidance is added, performance is essentially unchanged.

As motion is best seen in videos, we also include videos of how the symmetry loss and body shape conditioning affect motion quality; please see the supplementary webpage.

\begin{table}
\begin{center}
\scalebox{0.85}{
\begin{tabular}{lll c c}
\toprule
& & & \textbf{Fail} & \textbf{Traj Follow}  \\
\textbf{Terrain} & \textbf{Model} & \textbf{Guide} & Rate & Error  \\
\midrule
Flat & Agent Unaware & None &  0.252 & 0.102  \\
(Crowd) & Agent Aware & None &  0.087 & 0.082  \\
       & Agent Unaware & Agt Avoid & 0.060 & \textbf{0.067}  \\
       & Agent Aware & Agt Avoid &  \textbf{0.013} & 0.071   \\
\midrule
Random  & Body Unaware & None & 0.125 & 0.105  \\
         & Body Aware & None & \textbf{0.093} & \textbf{0.104}  \\
         & Body Unaware & Waypoint & \textbf{0.103} & \textbf{0.102} \\
         & Body Aware & Waypoint & 0.107 & 0.111  \\
\bottomrule
\end{tabular}}
\end{center}
\vspace{-6mm}
\caption{\animname ablation study while using \name as the trajectory planner.
}
\label{tab:supp-pacer-ablation}
\vspace{-3mm}
\end{table}
\begin{figure}[t]
\begin{center}
\includegraphics[width=\linewidth]{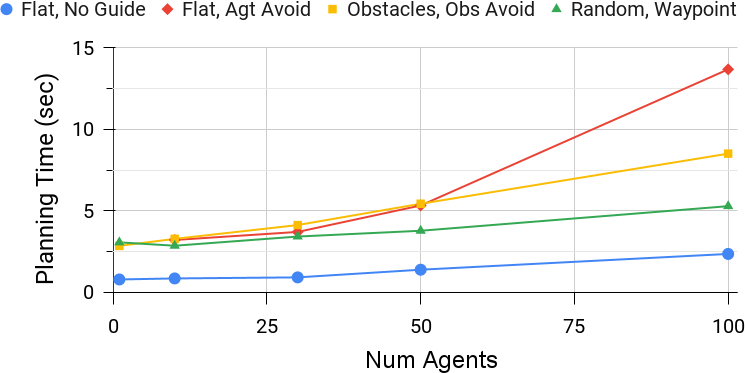}
\vspace{-8mm}
\end{center}
  \caption{TRACE planning time within the end-to-end pedestrian animation system for varying terrains, guidance, and number of simulated agents.
  }
\vspace{-2mm}
\label{fig:supp-trace-runtime}
\end{figure}

\subsection{Runtime Analysis}

\cref{fig:supp-trace-runtime} shows an analysis of the average runtime for one planning step of TRACE within the end-to-end animation system (on an NVIDIA TITAN RTX). Varying numbers of simulated humanoids are tested using the terrains and guidance introduced in Sec 4.3 of the main paper. 
With $\leq$50 agents, TRACE planning takes $\leq$5 sec, but becomes more costly with 100 agents, especially using agent avoidance guidance. Since collision avoidance requires pairwise comparisons between many agents, it can be costly.

The standalone PACER model is real time, running at $\sim$30 fps for 1 humanoid and $\sim$25 fps for 100.

\section{Discussions and Limitations}
\label{sec:supp-discussion}

\paragraph{\name Efficiency}
The main limitation of using our system in a real-time setting is the speed of the denoising process.
This is a well-known issue with diffusion models, and the community is actively working to address it.
For example, recent work on distilling diffusion models~\cite{meng2022distillation} could be applied here to greatly speed up sampling.

\paragraph{Multi-Objective Guidance}
One challenge with using several objectives simultaneously to guide \name is balancing the weight $\alpha$ for each. 
Though it is not difficult to tune each weight individually, we found that when combined, the guidance strength can be too much depending on the scene. Intuitively, if two guidance objectives are pushing a trajectory in the same direction (\eg avoiding obstacle collision and going to a waypoint), the combined guidance will have compounded strength that may push the trajectory to diverge off-manifold.
Work in image generation has noticed similar effects when using strong guidance, which manifests itself as saturated images. To avoid this, various forms of dynamic clipping during sampling have been introduced~\cite{saharia2022photorealistic}.
While this makes sense for images that have been normalized in a fixed range, it is not trivial for trajectories and we think this is an interesting problem for future work.

\paragraph{PACER Motions}
Though PACER is robust and traverses diverse terrains while driving humanoids with different body shapes, it struggles with large obstacles when there is no way around them. The motion generated at low speed can also be unnatural as our motion database contains few samples where the humanoid is traveling at extremely low speed. Our humanoids also lack motion diversity, since most body types will have similar walking gaits and will not manifest common pedestrian behaviors such as talking on the phone or with each other. More research is needed to improve the quality and diversity of the motion.

\end{document}